\pdfoutput=1

\documentclass[11pt]{article}

\usepackage[preprint]{acl}

\usepackage[ruled,vlined]{algorithm2e}

\usepackage{enumitem} 
\usepackage{times}
\usepackage{latexsym}
\usepackage{multirow}
\usepackage{amsmath}
\usepackage{booktabs}
\usepackage{graphicx}      
\usepackage{pifont}
\usepackage{stfloats}              
\usepackage{pdfpages}

\usepackage{subcaption}

\usepackage[T1]{fontenc}

\usepackage[utf8]{inputenc}

\usepackage{listings}
\usepackage{microtype}

\usepackage{inconsolata}

\usepackage{booktabs}
\usepackage{tabularx}   
\usepackage{array}
\usepackage{ragged2e}   
\usepackage{fontawesome5}

\newcolumntype{P}[1]{>{\RaggedRight\arraybackslash}p{#1}}

\title{\textsc{MoMentS}: A Comprehensive Multimodal Benchmark for Theory of Mind}

\author{
 \textbf{Emilio Villa-Cueva\textsuperscript{1}},
 \textbf{S M Masrur Ahmed\textsuperscript{2}},
 \textbf{Rendi Chevi\textsuperscript{1}},
 \\
 \textbf{Jan Christian Blaise Cruz\textsuperscript{1}},
 \textbf{Kareem Elzeky\textsuperscript{1}},
 \textbf{Fermin Cristobal\textsuperscript{1}},
 \\
 \textbf{Alham Fikri Aji\textsuperscript{1}},
 \textbf{Skyler Wang\textsuperscript{3}},
 \textbf{Rada Mihalcea\textsuperscript{4}},
 \textbf{Thamar Solorio\textsuperscript{1}}
\\[0.2em]
 \textsuperscript{1}MBZUAI,
 \textsuperscript{2}University of Houston,
 \textsuperscript{3}McGill University,
 \textsuperscript{4}University of Michigan
 \\[0.5em]
\faGithub \hspace{0.1em} \texttt{\href{https://github.com/villacu/MoMentS}{github.com/villacu/MoMentS}}
}

\begin{document}
\maketitle

\begin{figure*}[b!]
  \centering
  \includegraphics[trim={0cm 0.25cm 0cm 0.5cm},clip,width=1.0\linewidth]{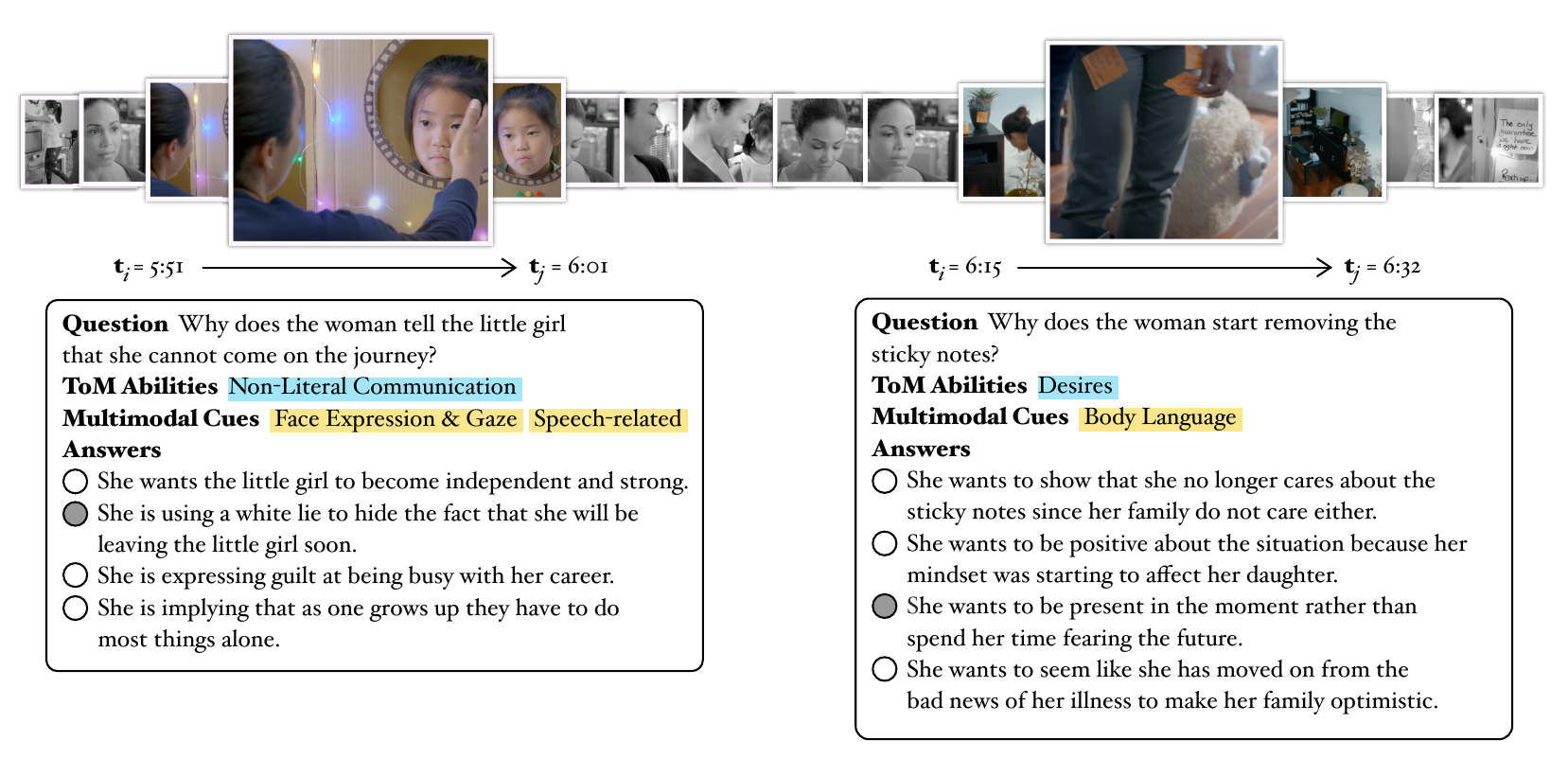}
  \caption{Overview of \textsc{MoMentS} questions.}
  \label{fig:teaser}
\end{figure*}

\begin{abstract}
Understanding Theory of Mind is essential for building socially intelligent multimodal agents capable of perceiving and interpreting human behavior. We introduce \textsc{MoMentS} (Multi\textbf{mo}dal \textbf{Ment}al \textbf{S}tates), a comprehensive benchmark designed to assess the ToM capabilities of multimodal large language models (LLMs) through realistic, narrative-rich scenarios presented in short films. \textsc{MoMentS} includes over 2,300 multiple-choice questions spanning seven distinct ToM categories. The benchmark features long video context windows and realistic social interactions that provide deeper insight into characters’ mental states. 
We evaluate several MLLMs and find that although vision generally improves performance, models still struggle to integrate it effectively. For audio, models that process dialogues as audio do not consistently outperform transcript-based inputs. Our findings highlight the need to improve multimodal integration and point to open challenges that must be addressed to advance AI’s social understanding.

\end{abstract}

\section{Introduction}

Throughout our lives, we continuously generate hypotheses about other people's emotions, knowledge, and a range of other mental states; these hypotheses guide how we understand and interact with others. This ability, known as Theory of Mind (ToM) \cite{premack1978does}, is essential for interpreting behavior at the individual level and fundamental to coherent human social interaction \cite{byom2013theory}.

Humans rely on more than just language to express their mental states. Gaze, facial expressions, body posture, gestures, and vocal cues all play an important role in communicating how we feel and what we think. This combination of verbal and non-verbal cues provides relevant multimodal information to infer mental states of others \cite{byom2013theory, bayliss2006predictive, desoneville2002facial}. 

For artificial agents, this information can serve as multimodal input that enhances socially intelligent behavior, empowering users across a wide range of applications: from facilitating communication and enhancing collaboration to offering companionship. A robust ToM enables such systems to anticipate intentions, understand desires and emotions, and detect knowledge gaps, to adapt their behavior to support users more effectively \cite{oguntola2021deepinterpretablemodelstheory}. Importantly, this requires not only inferring individual mental states, but doing so in context—accurately "reading the room" by processing these signals to interpret human behavior in socially situated settings \cite{williams2022supporting}.

Most existing benchmarks proposed to measure ToM in artificial agents predominantly center around belief-tracking tasks within text-based narratives or simplified multimodal settings \cite{chen2025theory}. While these approaches evaluate models' ability to reason about \textit{who knows or believes what}, they frequently neglect the interplay of emotions, intentions, pragmatic communication, and social contexts that characterize genuine human interactions. Consequently, a clear gap exists between existing evaluations and the richer, socially grounded reasoning required in realistic scenarios.

To support the development of socially intelligent multimodal agents and assess current models' ToM in realistic, socially grounded scenarios, we introduce \textbf{\textsc{MoMentS} (Multimodal Mental States)}, a comprehensive multimodal video question-answering benchmark designed to evaluate ToM across seven abilities derived from the ATOMS taxonomy \cite{beaudoin2020systematic}: Intentions, Desires, Beliefs, Knowledge, Percepts, Non-literal Communication, and Emotions. The dataset comprises 2,335 \textit{human-annotated} questions and 9,340 candidate answers sourced from 168 long-form videos, annotated with short and long context windows, multimodal cue markers, and adversarially-generated distractors to minimize biases.

To the best of our knowledge, \textsc{MoMentS} is the first benchmark to evaluate multimodal ToM in real-world videos with human actors, framing it explicitly as a socially situated ability. Our contributions are as follows:

 \begin{itemize}
 \item \textbf{\textsc{MoMentS}}: A novel multimodal benchmark with over 2,300 questions from real-world, long-form video data, explicitly structured to assess diverse ToM abilities.
 \item An \textbf{LLM-in-the-loop annotation framework} designed to produce challenging distractors and mitigate bias in answer sets.
 \item A baseline evaluation of multimodal LLMs, highlighting that although visual information improves performance, current models still predominantly rely on textual cues, underscoring the need for improved multimodal integration throughout the reasoning process.
 \end{itemize}
 
\section{Related Work}

Prior benchmarks for ToM broadly fall into two categories: text-only and multimodal. Traditional text-only benchmarks, such as TOMI \cite{le2019revisiting} and HI-TOM \cite{he2023hi}, predominantly focus on probing models' ability for nested belief tracking and logical inference through text stories lacking realistic social context. TOMBench \cite{chen2024tombench} expands beyond belief tracking alone, incorporating a broader taxonomy of social and pragmatic ToM tasks (e.g., faux-pas detection, persuasion, hidden emotions, desires) within everyday textual scenarios. Despite this richer coverage, it remains constrained by its purely textual format, lacking multimodal information critical to human social understanding \cite{byom2013theory}.

Multimodal approaches such as MMToM-QA \cite{jin2024mmtom} present procedurally-generated videos of single actors in household tasks, primarily evaluating goal and belief inferences without meaningful social interaction or emotional complexity. Similar to the text-only evaluations discussed above, this setup fails to reflect the depth and nuance of genuine human social behavior, limiting its applicability in evaluating socially intelligent AI systems \cite{chen2025theory}. 

From the social intelligence perspective, Social IQa \cite{sap2019socialiqa} probes social and emotional intelligence of models through multiple choice questions that require reasoning about social motivations, reactions, and actions based on specific situations. SOTOPIA \cite{zhou2023sotopia} evaluates how models navigate complex social scenarios and achieve social goals. EmoBench \cite{sabour2024emobench} measures emotional intelligence by assessing models’ ability to understand and apply emotional knowledge in complex social scenarios. However, these works are again limited to text-only evaluations and do not measure ToM directly. 

Social Genome \cite{mathur2025socialgenomegroundedsocial} (based on SocialIQ2 \cite{siq2}) addresses the evaluation of social interaction understanding in VLMs through video-based multiple-choice questions, but videos are limited to 60 second clips, and evaluation is not designed to evaluate ToM. Furthermore, \citet{guo-etal-2023-desiq} observed a strong bias in the representations of correct and incorrect answer candidates, where LLMs can achieve high accuracy with no context at all.

Given the limitations in prior work, there is a need for evaluating ToM within realistic multimodal settings, capturing authentic social interactions beyond goals and beliefs alone \cite{chen2025theory}.

\section{Dataset Design}

Recognizing the limitations of previous benchmarks, we design \textsc{MoMentS} based on two core principles: (1) an \textit{established taxonomy} of socially relevant ToM abilities --Emotions, Non-Literal Communication, Desires, Intentions, Knowledge, Percepts, and Beliefs-- to evaluate ToM beyond the commonly addressed belief and goal probing abilities, and (2) \textit{long-form videos} with real human actors that provide sufficient context and multimodal signals (e.g., facial expressions, gaze, body language, speech tone) to characterize interpersonal dynamics and mental states. This section outlines our taxonomy for probing different ToM abilities, the video selection process, and the annotations included in each question.

\begin{figure*}
  \centering
  \includegraphics[trim={0cm 0cm 0cm 0cm},clip,width=1.0\linewidth]{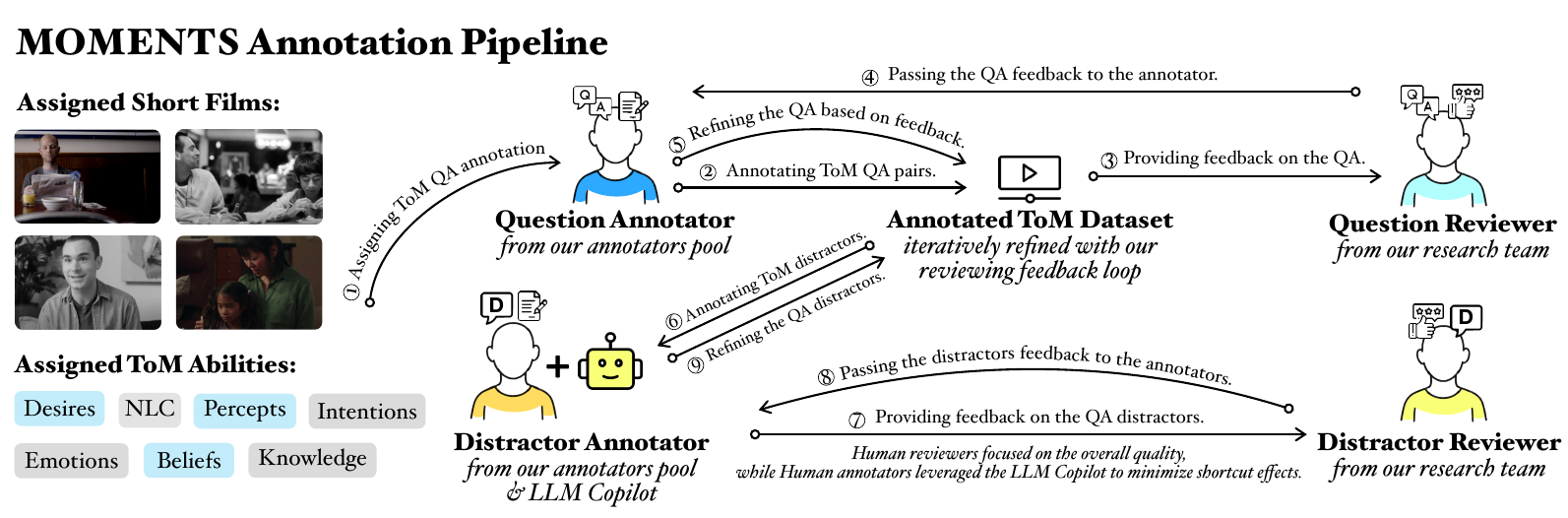}
  \caption{\textsc{MoMentS} Annotation Pipeline. Different colored t-shirts represent different annotators/reviewers.}
  \label{fig:pipeline}
\end{figure*}

\subsection{Taxonomy and Question Design}
\label{sec:atoms}
We adopt the \textbf{ATOMS taxonomy} (Abilities in Theory of Mind Space) introduced by \citet{beaudoin2020systematic} from their meta-analysis of ToM studies and proposed as a systematic framework for model evaluation by \citet{ma-etal-2023-towards-holistic}.
ATOMS categorizes ToM into seven distinct abilities: Knowledge, Emotions, Desires, Beliefs, Intentions, and Non-literal Communication (NLC). We describe and exemplify each ability in Table \ref{tab:tom-abilities}. This taxonomy supports precise question formulation and provides a detailed framework for systematically evaluating specific ToM abilities in models. We design annotation guidelines (See Appendix \ref{sec:guidelines}) around it. 


\begin{table*}[ht]
\footnotesize
\setlength{\tabcolsep}{4pt}
\renewcommand{\arraystretch}{1.1}
\centering
\begin{tabularx}{\linewidth}{@{}l
                                   P{0.32\linewidth}  
                                   P{0.23\linewidth}
                                   P{0.23\linewidth}@{}}
\toprule
\textbf{ToM Ability} & \textbf{Overview} & \textbf{Example Q} & \textbf{Example A}\\
\midrule
Knowledge &
Understanding what a person knows or does not know based on their sensory access. &
Why is the soldier interested in the boy's bottle? &
The soldier does not know what is inside the bottle and wants to find out. \\
\midrule
Emotions &
Identifying and reasoning about emotional responses, their evolution, and when emotions are hidden or complex. &
How do the old woman and the young woman feel in this conversation? &
The younger woman feels annoyed, and the older woman feels angry.\\
\midrule
Desires &
Situations that involve preferences, conflicting desires, or actions driven by desire. &
What does the girl want after walking past the group and reading the sign? &
She wants to go in to the establishment, finding it appealing. \\
\midrule
Beliefs &
Comprehending true and false beliefs and how beliefs influence actions. &
What does the woman with the ponytail think of the man who is watching TV? &
She thinks the man watching TV is aggressive.\\
\midrule
Intentions &
Understanding goals, motivations, and the underlying reasons for actions. &
Why does the old man give a beer to the bearded man and leave the cabin? &
He wants the bearded man to follow him so they can talk outside. \\
\midrule
Percepts &
Reasoning about what a character can or cannot perceive through their senses. &
Why didn't the woman with the long hair protect herself from the man? &
Because the man came up behind her and she didn't see him.\\
\midrule
NLC &
Interpreting humour, sarcasm, deception, and other speech that goes beyond literal meaning. &
Why does the young man in white ask the man in blue if he likes his work? &
He's being sarcastic and wants to annoy the man in blue. \\
\bottomrule
\end{tabularx}
\caption{Overview for ATOMS abilities with example question/answer pairs extracted from \textsc{MoMentS}}
\label{tab:tom-abilities}
\end{table*}

\subsection{Video Selection}

Existing datasets contain synthetic videos or minute-long clips that provide short temporal context. We instead propose to use \textit{short films} as these contain more complex characterizations and longer temporal contexts, while having a self-contained narrative. Our videos come from the SF20K dataset \cite{ghermi2025longstoryshortstorylevel}, which contains a curated collection of short films from the YouTube channel \textit{Omeleto}. \citet{ghermi2025longstoryshortstorylevel} verified that these films exhibit minimal information leakage to state-of-the-art language models compared to other common video sources like the sitcom \textit{Friends}. Additionally, the videos are high-quality, vary in length (10 to 20 minutes), and provide complete, cohesive stories.

Not all short films have scenarios suitable for evaluating ToM. To filter these out, we prompted GPT-4o with film synopses to identify ones that likely contain interesting social dynamics. We then select videos with the highest likelihood of generating meaningful question-answer pairs and assign each annotator a subset of these to annotate.

\subsection{Data Organization}
\label{sec:data_organization}

In line with prior work, we adopt a multiple-choice question-answer (MCQA) format, where each question includes one correct answer and three plausible but incorrect distractors. Figure \ref{fig:teaser} exemplifies two items from \textsc{MoMentS}, and more representative examples are presented in Appendix \ref{sec:moments_examples}. Below, we describe the structure and annotations included in each data point:

\paragraph{Questions} are derived from specific scenes in the short films and must probe one or more ToM abilities as defined in the ATOMS taxonomy. 

\paragraph{Answer Set} includes one correct option and three distractors. Annotators are instructed to write distractors that are as plausible as possible, such that only a nuanced understanding of the context can lead to the correct answer. We paid special attention to the distractors, see Section \ref{sec:copilot} for more details on this.

\paragraph{Tags for ToM Abilities} specify which ToM abilities (See Table \ref{tab:tom-abilities}) are targeted by the question. Questions may be annotated with multiple abilities, acknowledging that these often intersect in various scenarios.

\paragraph{Timestamps} mark the start and end of the video segment relevant to the question. Each question is annotated with two context windows:

\begin{itemize}
\item \textbf{Full Context Window} \boldmath$[t_{0}, t_{j}]$\unboldmath: A longer segment starting from the beginning of the video, intended to provide full narrative context useful for understanding character backgrounds, motivations, and evolving social dynamics.
\item \textbf{Focused Context Window} \boldmath$[t_i, t_j]$\unboldmath: A shorter segment containing only the immediate context required to answer the question. This window excludes broader narrative information, focusing instead on the specific scene being queried.

\end{itemize}

During evaluation, we explicitly instruct models that the question refers to the end of the provided interval ($t_j$). This approach minimizes reliance on temporal references in the question that may hint at the correct answer, which requires understanding the interaction. If leveraged effectively, the Full Context Window provides all the information required to understand characters, providing better insights into their mental states and interpersonal dynamics. 

\paragraph{Multimodal Cue Tags} indicate whether answering the question relies on interpreting specific nonverbal or auditory signals. These tags were optionally marked by annotators and are present only when such cues were deemed necessary for understanding the interaction. The possible cues include: \textit{"Facial Expressions or Gaze"}, \textit{"Body Language"}, and \textit{"Speech-related"}.


\section{Annotation Methodology}
\label{sec:annotation}

Creating multiple-choice questions for this task is challenging. Annotators must understand different ToM abilities, find relevant moments in short films, and write clear questions. Making good distractors is also difficult because humans often create distractors that models can easily guess without seeing the video context, as observed by \citet{guo-etal-2023-desiq} in other multimodal social understanding datasets. 

To address these challenges, we conducted two pilot annotation rounds (see Appendix \ref{sec:pilot}) before launching the main annotation phase. Findings from the pilots helped us refine our pipeline to address the cognitive demands of ToM question creation, reduce annotation biases, and ensure question quality. The final methodology included carefully structured annotation phases, refined guidelines, and a custom-built platform to support robust distractor generation.

\subsection{Annotation Pipeline}

\begin{table}[]
\centering
\resizebox{0.7\columnwidth}{!}{%
\begin{tabular}{ll}
\toprule
\textbf{Statistic} & \textbf{Length} \\
\midrule
Question Length & \(12.64\pm 4.2\) \\
Correct Answer Length & \(14.62 \pm 7.8\) \\
Distractor Length & \(14.97 \pm 7.7\) \\
$[t_i,t_j]$ length (s) & \(42.44 \pm 55.5\) \\
$[t_0,t_j]$ length (s) & \(388.47 \pm 262.3\)\\
\midrule
Number of Videos & \(168\) \\
Video length (m) & \(14.56 \pm 4.65\)\\
\bottomrule
\end{tabular}%
}
\caption{\textbf{Top:} Mean length $\pm$ SD of questions, correct answers, and distractors (in words), together with the average duration of the Focused $[t_i,t_j]$ and Full $[t_0,t_j]$ Context Windows (in seconds). \textbf{Bottom:} Number of Videos and average duration (in minutes)}
\label{tab:statistics}
\end{table}

Annotation guidelines were centered around the ATOMS taxonomy and the specific goals of the benchmark. They included illustrative examples, key indicators (\textit{what to look for}) for each ToM ability, and clearly defined criteria for both acceptable and problematic question types. We iteratively refined the guidelines based on feedback from our expert sociologist and from the annotators themselves during the pilot runs.

The main annotation phase spanned six weeks and involved 16 annotators who collectively produced 2,335 questions. This phase followed the methodology developed during the second pilot and incorporated several design choices aimed at improving quality and reducing bias (see Figure \ref{fig:pipeline} for an overview):

\begin{itemize}[nosep,leftmargin=1.2em,labelsep=0.4em]
\item Annotators were asked to watch the full short film before writing questions to ensure understanding of character motivations and social dynamics.
\item Each was assigned 2--3 ToM abilities to specialize in, promoting category-specific expertise.
\item The schedule alternated weekly: a week focused on writing questions, the next on creating distractors for peers' questions. During the distractor-creation stage, annotators flagged poorly written or overly subjective questions, adding a layer of peer-based quality control.
\item A custom platform integrated an LLM for real-time distractor feedback, flagging biased sets automatically (Section~\ref{sec:copilot}).
\item We provided weekly feedback based on a review of the submitted material. For questions, we emphasized clarity, appropriate ToM category assignment, and avoidance of overly subjective QA pairs. For distractors, we focused on ensuring that none of the distractors could be considered a "technically correct" answer.
\item We provided bonuses for early submissions and for the annotators who produced the highest-quality questions.
\end{itemize}
This approach encouraged focused annotation, peer-based quality control, and robust distractor generation, resulting in the final \textsc{MoMentS} evaluation dataset. Table \ref{tab:statistics} report statistics about the dataset. In Appendix \ref{sec:data_stats}, we report the demographics of annotators, cost of the annotations, and number of questions associated to each ToM ability.

\subsection{Framework for Distractor Creation}
\label{sec:copilot}

Models frequently rely on subtle biases to guess correctly; our initial pilot batch showed this issue with models consistently achieving non-trivial performance by identifying correct answers without the required context. Creating high-quality distractors remains challenging for annotators despite providing them with guidelines; even subsequent re-annotation of distractors by us similarly demonstrated persistent biases.

While various post-hoc strategies exist to mitigate distractor bias \cite{shortcuts,guo-etal-2023-desiq}, we integrate bias prevention directly into the annotation workflow. We designed a custom annotation platform embedded with an LLM acting as an on-the-fly evaluator for newly proposed distractor sets.

Given a question with one correct answer and three proposed distractors, the platform evaluates potential biases distractor as described in Algorithm \ref{alg:distractor_eval}.

\begin{algorithm}[!htb]
\centering
\caption{Distractor Set Assessment}
\label{alg:distractor_eval}
\resizebox{0.8\linewidth}{!}{%
\begin{minipage}{\linewidth}
\KwIn{Question $Q$, correct answer $a^*$, distractors $D=\{d_1,d_2,d_3\}$, trials $N$, threshold $k$.}
\KwOut{Indicator of biased distractors}
\BlankLine
$c \leftarrow 0$\;
\For{$i \leftarrow 1$ \KwTo $N$}{
  $A \leftarrow \text{shuffle}(\{a^*\} \cup D)$\;
  $a \leftarrow \text{LLMAnswer}(Q, A)$\;
  \If{$a = a^*$}{
    $c \leftarrow c + 1$\;
  }
}
\If{$c \geq k$}{
  \Return \textbf{flag biased}\;
}
\end{minipage}%
}
\end{algorithm}

We establish empirically determined $k = 5$ and $N = 6$ to balance reliability and computational efficiency. A distractor set is flagged as biased if the model identifies the correct answer $k$ or more times out of $N$ trials. We initially employed GPT-4o-mini as the LLM for the first 800 questions; as we observed that the cost was relatively low, we decided to use GPT-4o for the remaining annotations.

\section{Experimental Evaluations}

We conduct experiments to evaluate the performance of current multimodal models in inferring mental states and to identify the factors that influence their performance. Specifically, we aim to answer: (i) How well do these models perform overall and across different ToM abilities? (ii) To what degree does visual information and context length impact performance? and (iii) How effective is our LLM-in-the-loop distractor creation platform at mitigating answer set biases? To this end, we report model accuracies on \textsc{MoMentS}, ablate the effect of the visual modality and context window length, and assess performance in a no-context setting against baselines lacking bias-mitigation mechanisms.

\subsection{Experimental Setup}

\begin{table}[h]
\centering
\resizebox{\columnwidth}{!}{%
\begin{tabular}{lllll}
\toprule
 Model                 & \multicolumn{2}{c}{$[t_0,t_j]$}                       & \multicolumn{2}{c}{$[t_i,t_j]$}                   \\
                  \midrule
 & \multicolumn{1}{c}{$T$} & \multicolumn{1}{c}{$VT$} & \multicolumn{1}{c}{$T$} & \multicolumn{1}{c}{$VT$} \\
\midrule
\ding{108} LLaVA-Video-7B    & 47.0                  & 49.36 (+2.3)           & 45.6                  & \textbf{52.01} (+6.4)           \\
\ding{108} InternVL2.5 8B    & 46.0                  & 46.63 (+0.6)           & 45.4                  & 51.79 (+6.4)           \\
\ding{108} LongVA-7B-DPO     & 41.0                  & 44.24 (+3.3)           & 41.5                  & 44.5 (+3)              \\
\ding{108} Qwen2.5 VL 7B     & 41.3                  & 38.05 (+-3.3)          & 38.4                  & 44.33 (+5.9)           \\
\midrule
\ding{108} LLaVA-Video-72B   & 63.2                  & 65.96 (+2.8)           & 62.1                  & \textbf{67.66} (+5.6)           \\
\ding{108} InternVL2.5 78B   & 53.3              &   61.09 (+7.8)            & 52.2                  & 61.48 (+9.3)           \\
\midrule
                  & \multicolumn{1}{c}{$T$} & \multicolumn{1}{c}{$VT$} & \multicolumn{1}{c}{$T$} & \multicolumn{1}{c}{$VT$} \\
\midrule
\ding{115} Qwen2.5-Omni-7B   & 46.8                  & 53.41 (+6.6)           & 45.8                  & \textbf{56.19} (+10.4)          \\
\ding{115} VideoLLaMA2-7B-AV & 37.6                  & 40.96 (+3.3)           & 38.4                  & 43.13 (+4.7)           \\
\ding{115} MiniCPM-o 2.6 (8B)         & 47.8                  &                      & 47.1                  & 50.17 (+3.1)           \\
\midrule
                  & \multicolumn{1}{c}{$A$} & \multicolumn{1}{c}{$VA$} & \multicolumn{1}{c}{$A$} & \multicolumn{1}{c}{$VA$} \\
\midrule
\ding{115} Qwen2.5-Omni-7B   & 44.41	& 52.69 (+8.3)          & 48.46                  & \textbf{55.59} (+7.1)          \\
\ding{115} VideoLLaMA2-7B-AV & 34.22                 & 42.32 (+8.1)           & 34.22                & 43.6 (+9.4)           \\
\ding{115} MiniCPM-o 2.6 (8B)         & 39.5                  &                      & 39.6                  & 48.25 (+8.6)           \\
\midrule
                  & \multicolumn{2}{c}{$A$}                          & \multicolumn{2}{c}{$A$}                          \\
                  \midrule
\ding{116} Kimi-Audio-7B     & \multicolumn{2}{c}{31.7}                       & \multicolumn{2}{c}{\textbf{48.6}}                       \\
\ding{116} Qwen2-Audio-7B    & \multicolumn{2}{c}{34.5}                       & \multicolumn{2}{c}{35.5}                       \\
\midrule
                  & \multicolumn{2}{c}{$VA$}                         & \multicolumn{1}{c}{}  &                        
                  \\
                  \midrule
*Human (average of 3)            & \multicolumn{2}{c}{86.33 $\pm$ 1.15}                         & \multicolumn{1}{c}{}  &                       \\
*Human (majority-vote)            & \multicolumn{2}{c}{91.0}                         & \multicolumn{1}{c}{}  &                       \\
\bottomrule
\end{tabular}%
}
\caption{Accuracy of different models on \textsc{MoMentS}, reported for both Full $[t_0, t_j]$ and Focused $[t_i, t_j]$ context windows. For Video LLMs (\ding{108}), we show performance with transcripts only (T) and video+transcripts (VT). For Audiovisual LLMs (\ding{115}), we additionally report audio (A) and video+audio (VA) inputs. For Speech LLMs (\ding{116}), we report performance using audio only (A). *Human evaluation is carried out in a sample of 100 randomly drawn items.}
\label{tab:global_acc}
\end{table}

We evaluate three types of multimodal LLMs: Video, Audiovisual, and Speech-based.
\paragraph{Video LLMs} These models process visual and text inputs. We evaluate LLaVA-Video 7B and 72B \cite{zhang2024videoinstructiontuningsynthetic}, InternVL 2.5 8B and 78B \cite{chen2025expandingperformanceboundariesopensource}, LongVA 7B \cite{zhang2024longcontexttransferlanguage}, and Qwen2.5 VL 7B \cite{bai2025qwen25vltechnicalreport}. Each model receives 64 uniformly sampled frames per question (see Appendix \ref{sec:frame_ablation} for frame count ablations).

Since Video LLMs process only vision and text, we transcribe dialogues using ASR through WhisperX using Whisper \textit{large-v2} as the backbone model (refer to Appendix \ref{sec:asr} for an evaluation on the performance of the ASR system).

\paragraph{Audiovisual LLMs} These models process visual, text, and audio inputs. We evaluate Qwen2.5-Omni 7B \cite{Qwen2.5-Omni}, VideoLLaMa2 7B \cite{damonlpsg2024videollama2}, and MiniCPM-o 2.6 (8B) \cite{yao2024minicpm}. We observed that MiniCPM-o 2.6 yielded errors in long-form videos; therefore, for this model we only report results on the Focused Context Window.

\paragraph{Speech LLMs} These models process audio and text inputs. We evaluate Kimi-Audio 7B \cite{kimiteam2025kimiaudiotechnicalreport} and Qwen2 Audio \cite{chu2024qwen2audiotechnicalreport}.

All models are evaluated using the \textit{transformers} library \cite{wolf2020huggingfacestransformersstateoftheartnatural} with temperature set to 0. Under 70B parameters models run on a single NVIDIA A100 GPU, whereas larger models (>70B) run on three NVIDIA H100 GPUs.

\subsection{LLM Evaluation}
\label{sec:llm_eval}

We evaluate models under different input conditions: textual or audio dialogues only (Transcripts ($T$) or Audio ($A$)), and combined inputs (Vision+Transcripts ($VT$) or Vision+Audio ($VA$)). Additionally, we compare model performance across two temporal contexts: the Full Context Window $[t_0,t_j]$ and the Focused Context Window $[t_i,t_j]$.

\paragraph{Global Accuracy}
Table \ref{tab:global_acc} reports the global accuracy on \textsc{MoMentS}; we observe that video input improves performance in most cases. However, the gains are modest, indicating that current models may underutilize visual cues. Performance tends to drop when using the longer Full Context Window, we attribute this to the fact that long video understanding is still challenging for open models.

As a reference, we conducted human evaluations on a subset of 100 randomly sampled questions with three different evaluators on the Full Context Window and the $VA$ setting. Individual accuracies were 87.0/85.0/87.0 (an average of 86.3). Majority-vote accuracy with ties marked as incorrect was 91.0. We observed a percent agreement of 0.80, and a Fleiss \(\kappa\) of 0.733, which indicates substantial agreement between evaluators when selecting an answer. In Table \ref{tab:global_acc}, we report both majority-vote (marking ties as incorrect) and the average of individual accuracies.

\begin{figure*}[t]                
  \centering
   \begin{subcaptionbox}{Boxplot of accuracies in the Focused Context Window.\label{fig:sub-b}}%
        [0.49\linewidth]
        {\includegraphics[width=\linewidth]{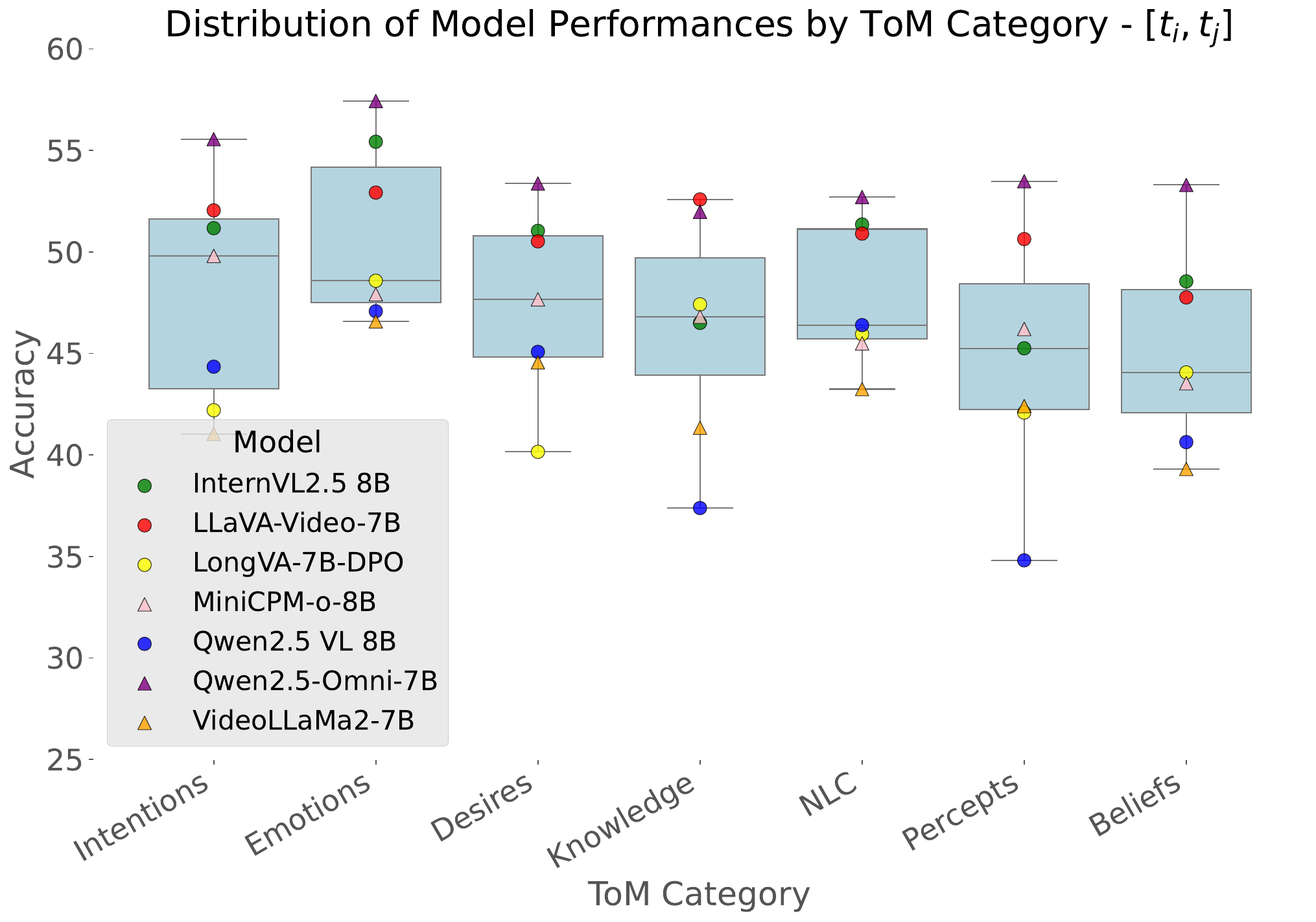}}
  \end{subcaptionbox}
  \hfill
  \begin{subcaptionbox}{Boxplot of accuracies in the Full Context Window.\label{fig:sub-a}}%
        [0.49\linewidth]                 
        {\includegraphics[width=\linewidth]{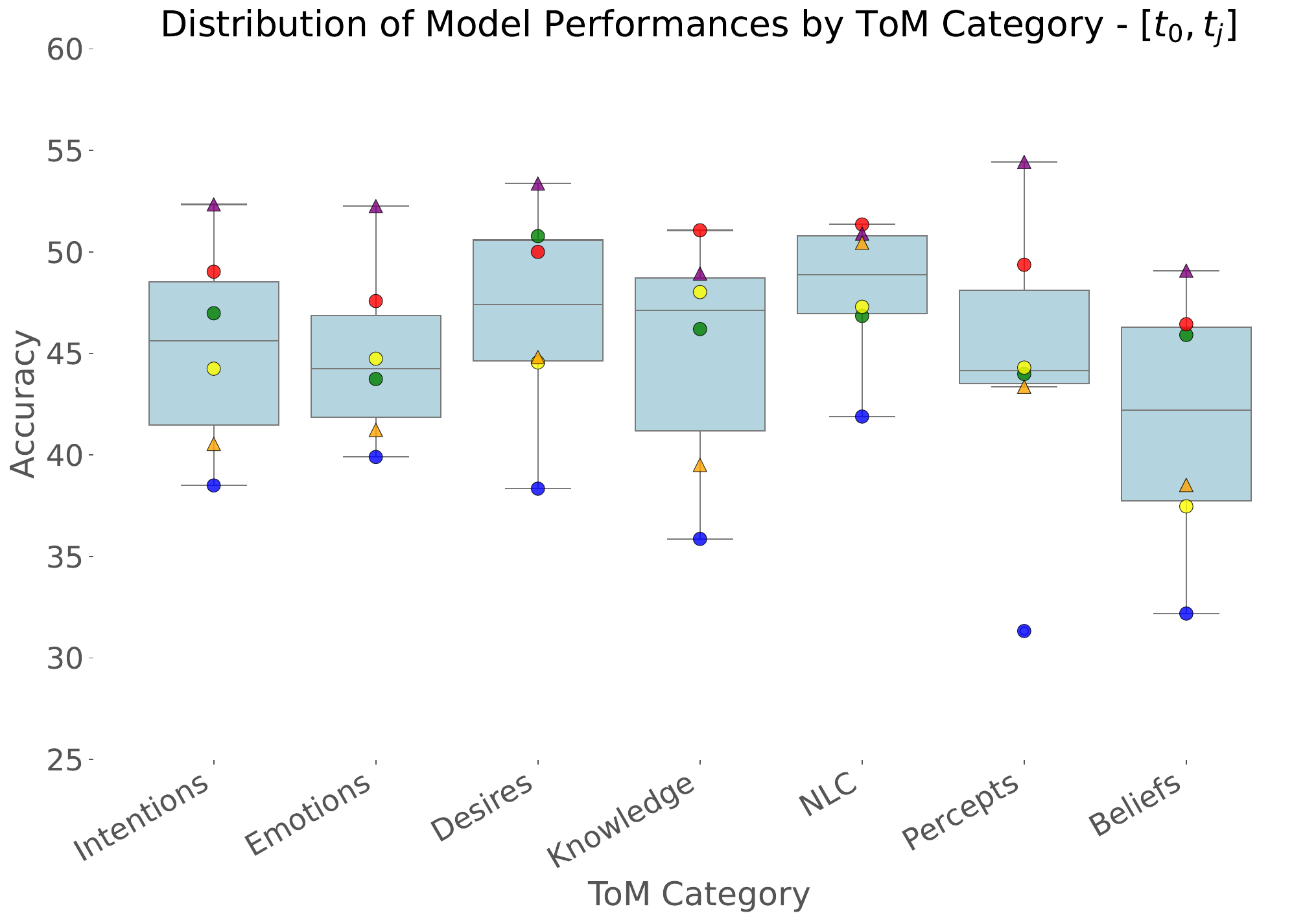}}
  \end{subcaptionbox}                              
 
  \caption{Boxplots comparing accuracies across different models (at $\sim$7B parameter scale) and ToM abilities. Results for Audiovisual LLMs (\ding{115}) are reported using video and audio inputs ($VA$), while results for Video LLMs (\ding{108}) use video and transcript inputs ($VT$).}
  \label{fig:boxplot}
\end{figure*}

\paragraph{Accuracy by ToM Ability}

Figure \ref{fig:boxplot} presents boxplots with per-model scatter points, showing model accuracy across different ToM abilities under two context window conditions. As observed in the global accuracies, models perform better using shorter Focused Context Windows, though the impact of context length varies by ability. For example, accuracy is notably higher for Emotions and Intentions questions with shorter contexts, suggesting that these tasks rely more on immediate cues. Among results within the longer Full Context Window, Knowledge, Desires, and Non-literal Communication (NLC) questions perform relatively better, suggesting that longer context may be beneficial for understanding characters' background and effectively answering these questions. For both of the context settings, Percepts and Beliefs remain the most challenging abilities. Future work should investigate how context window length affects human performance in this task.

\begin{figure*}[t]                
  \centering
   \begin{subcaptionbox}{Video LLMs ($\sim$7B scale)}%
        [0.3\linewidth]
        {\includegraphics[width=\linewidth]{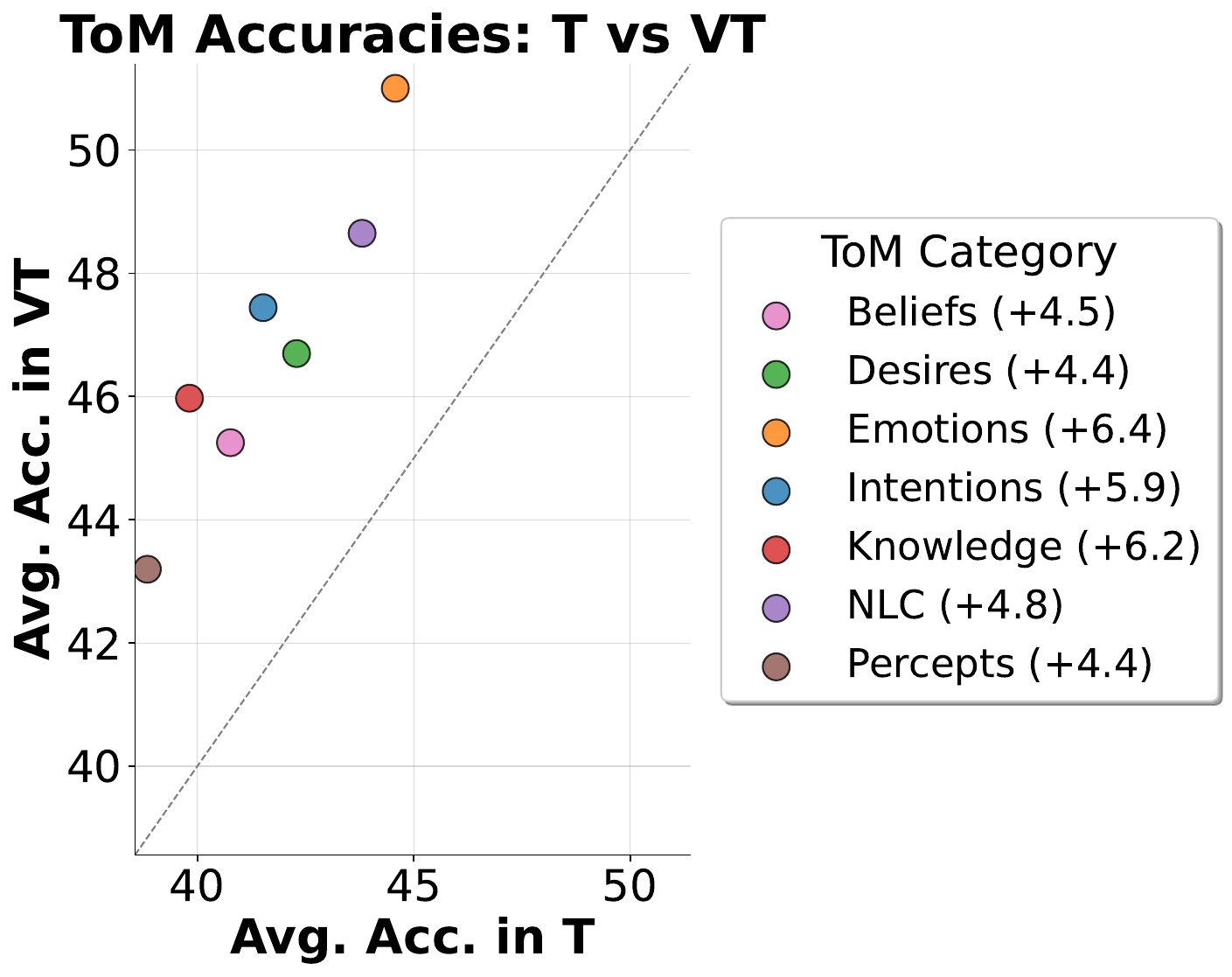}}
  \end{subcaptionbox}
  \hfill
  \begin{subcaptionbox}{Audiovisual LLMs ($VT$ vs. $T$)}%
        [0.3\linewidth]
        {\includegraphics[width=\linewidth]{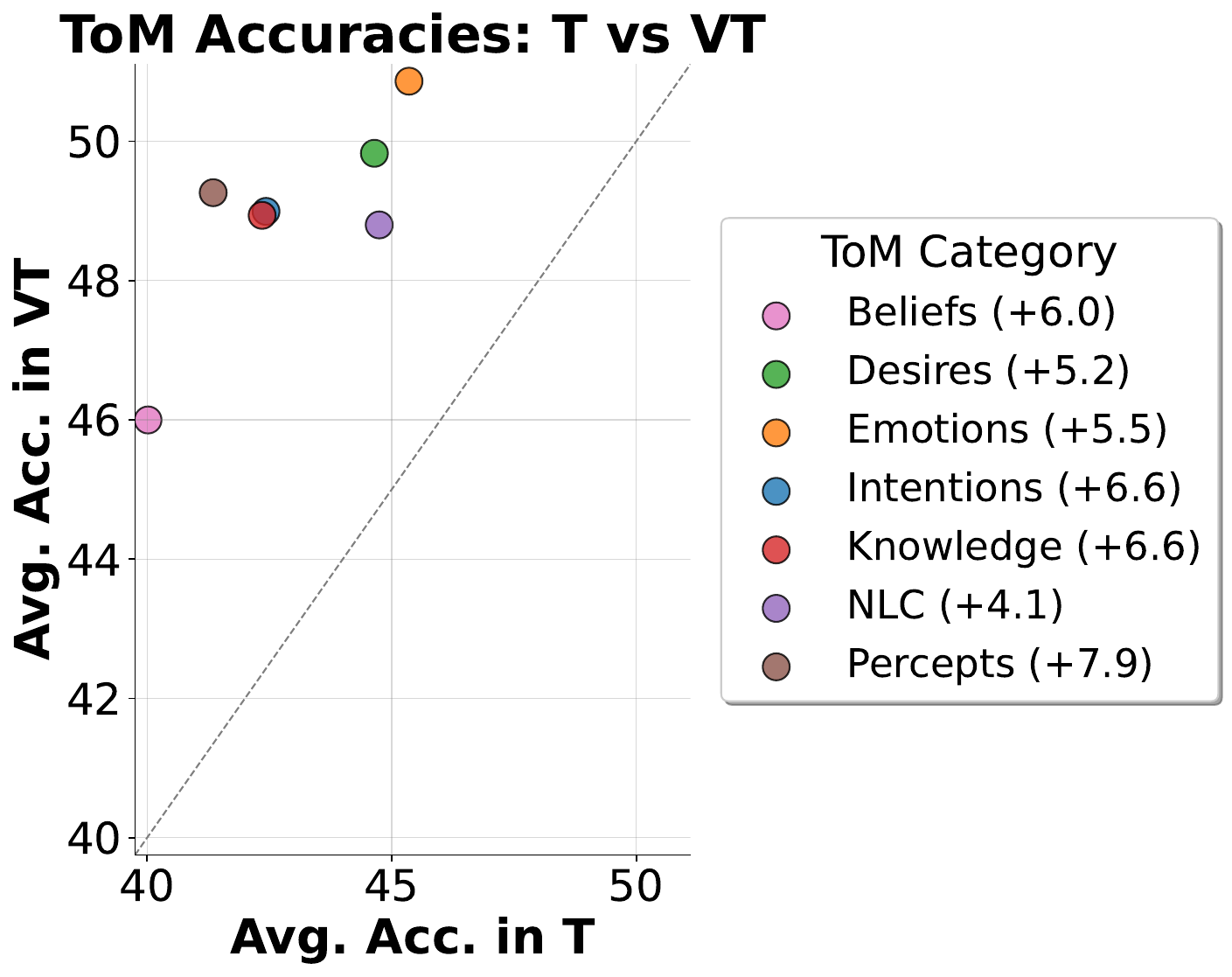}}
  \end{subcaptionbox}
  \hfill
  \begin{subcaptionbox}{Audiovisual LLMs ($VA$ vs. $A$)}%
        [0.3\linewidth]
        {\includegraphics[width=\linewidth]{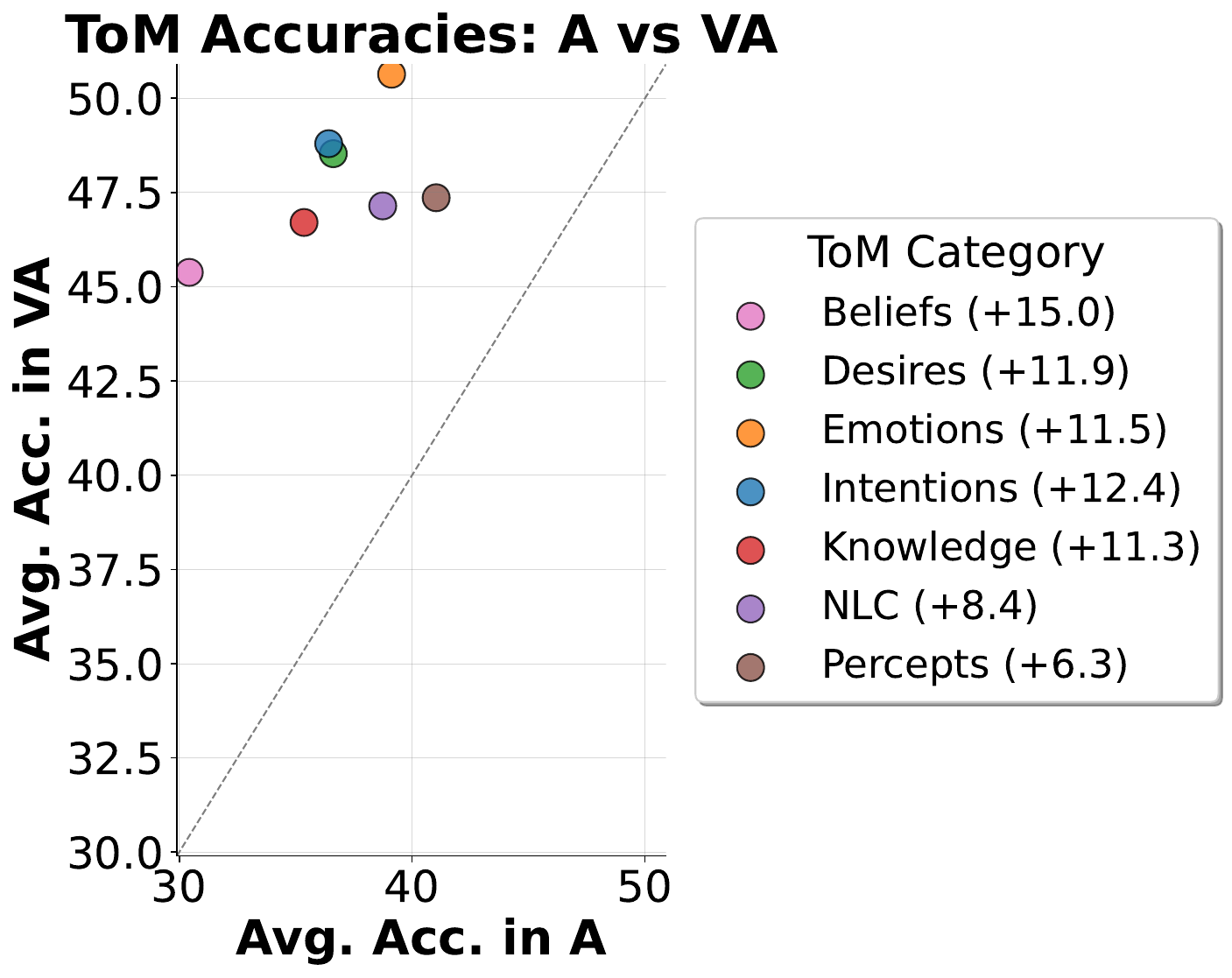}}
  \end{subcaptionbox}   
 
  \caption{Comparison between average accuracies across the evaluated Video LLMs and Audiovisual LLMs with and without vision across different ToM abilities (Focused Context Window). The number in parentheses refers to the improvement due to the visual modality.}
  \label{fig:vt_vs_t}
\end{figure*}

In Figure \ref{fig:vt_vs_t}, we compare the effect of visual input for Video and Audiovisual LLMs. Vision contributes positively across all abilities, although the improvement varies by model type. For Video LLMs, the gains are relatively consistent, while Audiovisual LLMs show greater variation, especially when dialogues are provided as audio: some abilities improve by over 12 points, others by only 6-8. This larger gain with vision is mainly due to the weaker performance of the audio-only setting, where models struggle on abilities such as Beliefs or Intentions that likely require more complex information that the visual channel can provide. Notably, Non-literal Communication shows the smallest improvement, suggesting stronger reliance on dialogue than visual cues compared to the other abilities.

\paragraph{Multimodal Cues}

\begin{table}[]
\centering
\resizebox{0.8\columnwidth}{!}{%
\begin{tabular}{llll}
\toprule
                 & $\Delta_{Focus-Full}$ & $\Delta_{VT-T}$ & $\Delta_{VA-A}$ \\
                 \midrule
Body Language    & 3.71    & 7.92                             & 8.05            \\
F. Exp. and Gaze & 2.01    & 6.5                              & 6.79            \\
Speech-related   & 1.77    & 3.62                             & 6.32          \\
\bottomrule
\end{tabular}%
}
\caption{Effect of context length and visual input on questions marked as reliant on multimodal cues. $\Delta_{Focus-Full}$ is the average accuracy difference between Focused and Full Contexts, with positive values indicating better performance in shorter intervals. $\Delta_{VT-T}$ and $\Delta_{VA-A}$ represent average accuracy gains from adding visual input (VT vs. T across Video and Audiovisual LLMs; VA vs. A across Audiovisual LLMs), both reported for the shorter context interval.}
\label{tab:social_signals}
\end{table}

We further analyze performance on questions requiring multimodal understanding (facial expression or gaze, body language, and speech-related cues). As shown in Table \ref{tab:social_signals}, incorporating visual input and using a shorter context window generally improves performance, particularly for questions involving Body Language and Facial Expressions or Gaze. Speech-related questions benefit less from visual input when dialogues are provided as transcripts ($VT$), confirming stronger reliance on text. Interestingly, Audiovisual models show larger gains in Speech-related questions when dialogues are provided as audio ($VA$), suggesting vision is more beneficial in this setting.

\subsection{Evaluation on Answer Set Bias}
In this section, we evaluate the effectiveness of using an LLM-in-the-loop design during the annotation pipeline, specifically for distractor creation. For MCQA-style ToM evaluation to be meaningful, questions should not be answerable without access to some form of context such as video, audio, or transcripts. However, as observed in our initial pilot and in prior work \cite{guo-etal-2023-desiq}, models often exploit biases in question-answer sets to guess the correct answer even without contextual input.

We assess the extent of this issue by comparing \textsc{MoMentS} to two baselines: our initial pilot (which did not use LLM assistance for distractor creation) and SocialIQ2, a similar video MCQA dataset. We prompt models with only the questions and answer options (without context) and measure their accuracy. As shown in Table~\ref{tab:biases}, our proposed LLM-assisted distractor generation substantially reduces answer-set bias and lowers model accuracy by over 20 percentage points, highlighting the effectiveness of our approach.

By reducing biases in the answer sets, we create greater headroom for models to improve through reasoning based on the provided context.

\begin{table}[]
\centering
\small
\resizebox{0.85\columnwidth}{!}{%
\begin{tabular}{llll}
\toprule
Model & SIQ2-dev & M-P1 & \textsc{MoMentS} \\
\midrule
Qwen2.5 VL 7B & 52.49 & 60.59 & 36.05 (-24.54) \\
LongVA-7B-DPO & 53.38 & 58.49 & 34.85 (-23.63) \\
LLaVA-Video-7B & 56.2 & 59.48 & 40.10 (-19.38) \\
InternVL2.5 8B & 51.43 & 55.39 & 36.26 (-19.13) \\
\bottomrule
\end{tabular}%
}
\caption{Accuracy by \textit{guessing} the correct answer, where models are not provided with any context about the question. M-P1 refers to our first pilot study, and SIQ2-dev to the development set of SocialIQ2 \cite{siq2}.}
\label{tab:biases}
\end{table}


\section{Open Challenges for Future Model Development}

Our evaluations on \textsc{MoMentS} suggest that current limitations in multimodal ToM performance may stem not only from the reasoning capabilities of large language models, but also from how these systems access and process multimodal evidence. Our findings point to several technical factors that likely limit models’ ability to reason about mental states in socially rich scenarios. In this section, we outline four open challenges that, if addressed, could foster progress toward building better social multimodal agents.

\paragraph{Capturing Prosody and Ambient Sound in Audio}
Transcripts alone omit environmental sounds and paralinguistic cues (speaker prosody, intonation), which support accurate inferences about \textit{Percepts}, \textit{Emotions}, \textit{Intentions}, and \textit{Non-literal Communication}. In addition, errors in the ASR propagate downstream. While some audio-native models like Kimi-Audio and Qwen2.5-Omni (audio-only setting, $A$) demonstrate slight advantages over transcripts when vision is not provided, most transcript-only models perform comparably or better when visual input is present (see Table \ref{tab:global_acc}). Future research should focus either on integrating rich audio descriptors into existing video-text pipelines or improving the current audio-processing capabilities of audiovisual models, especially for understanding longer conversational contexts.

\paragraph{Precise Vision–Speech Alignment}
Prior work has shown that state-of-the-art models struggle to attribute utterances to speakers in multimodal conversations \cite{chang2025multimodalconversationstructureunderstanding}. Answering \emph{Who said what, when?} requires time-synchronized links between each utterance, the speaking character, and the surrounding visual context. Without such alignment, models cannot track which speakers possess which knowledge, nor can they exploit gaze, facial expressions, or body language that modulate dialogue meaning. The small gains we observe from adding vision  (Table~\ref{tab:global_acc}), and the limited improvements on questions marked as reliant on visual cues (Table~\ref{tab:social_signals}), also indicate that existing pipelines underutilize this channel. 

\paragraph{Human-Centered Frame Selection}
Uniform frame sampling risks missing short yet meaningful signals while wasting computation on redundant content. Simply increasing the frame rate is expensive and, as our ablation in Appendix~\ref{sec:frame_ablation} shows, does not improve performance. Specialized frame sampling strategies that prioritize human-salient events (faces, hands, gaze shifts) are needed to capture the cues that observers actually rely on.

\paragraph{Structured Reasoning over Multimodal Evidence}
Reasoning improves a wide range of \emph{text-only} tasks, including ToM benchmarks. However, as \citet{mathur2025socialgenomegroundedsocial} reports, asking VLMs to reason neither boosts accuracy nor yields human-aligned explanations for social MCQA in videos. We argue that effective multimodal reasoning may be bottlenecked by the three challenges above: inadequate audio representations, weak vision–speech alignment, and sub-optimal frame selection. Until models receive richer, better-organized evidence, additional reasoning steps are unlikely to help.

\section{Conclusion}

We introduced \textsc{MoMentS}, a benchmark that probes seven ToM abilities in realistic, long-form videos. It contains over 2,300 human-annotated MCQA items with substantially reduced biases in answer sets compared to prior datasets. From baseline experiments with Video, Audiovisual, and Speech LLMs we observe: (i) visual input offers consistent yet modest gains, indicating underutilization of visual cues; (ii) using audio inputs does not yield a noticeable improvement over transcripts, suggesting that current models still have struggle to effectively integrate this modality; and (iii) performance tends to drop on extended context windows, highlighting limitations in long-range video reasoning. 

Based on these results, we identify several open challenges that likely constrain progress on multimodal ToM tasks, ranging from multimodal alignment and audio processing to frame selection and reasoning over multimodal evidence. Addressing these issues will be essential for developing AI systems capable of truly understanding, predicting, and responding to human mental states in real-world social settings.

\section*{Limitations}

We adopt a multiple-choice QA format in \textsc{MoMentS} to streamline annotation and ensure consistent evaluation. While this design supports scalable benchmarking, it limits analysis of lower-level behavioral cues such as turn-taking, speech acts, or gesture dynamics as we do not provide fine-grained annotations on them. Investigating the relation between these cues and specific ToM abilities remains an important direction for future work.
Additionally, \textsc{MoMentS} uses static video data, which does not capture model performance in interactive or dynamic social environments. Extending evaluation to such settings is a promising but currently challenging task, as it would require reliably simulating complex, multimodal human behaviors.
Finally, although using multiple annotators per question could reduce subjectivity, resource constraints limited us to one annotator per question. To mitigate this, we incorporated peer-checking during distractor creation and conducted multiple rounds of author review to ensure data quality and consistency.


\bibliography{custom_ai,custom_social}

\begin{thebibliography}{34}
\providecommand{\natexlab}[1]{#1}

\bibitem[{Bai et~al.(2025)Bai, Chen, Liu, Wang, Ge, Song, Dang, Wang, Wang, Tang, Zhong, Zhu, Yang, Li, Wan, Wang, Ding, Fu, Xu, Ye, Zhang, Xie, Cheng, Zhang, Yang, Xu, and Lin}]{bai2025qwen25vltechnicalreport}
Shuai Bai, Keqin Chen, Xuejing Liu, Jialin Wang, Wenbin Ge, Sibo Song, Kai Dang, Peng Wang, Shijie Wang, Jun Tang, Humen Zhong, Yuanzhi Zhu, Mingkun Yang, Zhaohai Li, Jianqiang Wan, Pengfei Wang, Wei Ding, Zheren Fu, Yiheng Xu, and 8 others. 2025.
\newblock \href {https://arxiv.org/abs/2502.13923} {Qwen2.5-vl technical report}.
\newblock \emph{Preprint}, arXiv:2502.13923.

\bibitem[{Bain et~al.(2023)Bain, Huh, Han, and Zisserman}]{whisperx}
Max Bain, Jaesung Huh, Tengda Han, and Andrew Zisserman. 2023.
\newblock \href {https://arxiv.org/abs/2303.00747} {Whisperx: Time-accurate speech transcription of long-form audio}.
\newblock \emph{Preprint}, arXiv:2303.00747.

\bibitem[{Bayliss and Tipper(2006)}]{bayliss2006predictive}
Andrew~P Bayliss and Steven~P Tipper. 2006.
\newblock Predictive gaze cues and personality judgments: Should eye trust you?
\newblock \emph{Psychological science}, 17(6):514--520.

\bibitem[{Beaudoin et~al.(2020)Beaudoin, Leblanc, Gagner, and Beauchamp}]{beaudoin2020systematic}
Cindy Beaudoin, {\'E}lizabel Leblanc, Charlotte Gagner, and Miriam~H Beauchamp. 2020.
\newblock Systematic review and inventory of theory of mind measures for young children.
\newblock \emph{Frontiers in psychology}, 10:2905.

\bibitem[{Bredin(2023)}]{pyannote}
Hervé Bredin. 2023.
\newblock {pyannote.audio 2.1 speaker diarization pipeline: principle, benchmark, and recipe}.
\newblock In \emph{Proc. INTERSPEECH 2023}.

\bibitem[{Byom and Mutlu(2013)}]{byom2013theory}
Lindsey~J Byom and Bilge Mutlu. 2013.
\newblock Theory of mind: Mechanisms, methods, and new directions.
\newblock \emph{Frontiers in human neuroscience}, 7:413.

\bibitem[{Chang et~al.(2025)Chang, Cramer, Ho, Nguyen, Yuan, and Bamman}]{chang2025multimodalconversationstructureunderstanding}
Kent~K. Chang, Mackenzie~Hanh Cramer, Anna Ho, Ti~Ti Nguyen, Yilin Yuan, and David Bamman. 2025.
\newblock \href {https://arxiv.org/abs/2505.17536} {Multimodal conversation structure understanding}.
\newblock \emph{Preprint}, arXiv:2505.17536.

\bibitem[{Chen et~al.(2025{\natexlab{a}})Chen, Jiang, Qin, and Tan}]{chen2025theory}
Ruirui Chen, Weifeng Jiang, Chengwei Qin, and Cheston Tan. 2025{\natexlab{a}}.
\newblock \href {https://doi.org/10.18653/v1/2025.acl-long.1522} {Theory of mind in large language models: Assessment and enhancement}.
\newblock In \emph{Proceedings of the 63rd Annual Meeting of the Association for Computational Linguistics (Volume 1: Long Papers)}, pages 31539--31558, Vienna, Austria. Association for Computational Linguistics.

\bibitem[{Chen et~al.(2025{\natexlab{b}})Chen, Wang, Cao, Liu, Gao, Cui, Zhu, Ye, Tian, Liu, Gu, Wang, Li, Ren, Chen, Luo, Wang, Jiang, Wang, He, Shi, Zhang, Lv, Wang, Shao, Chu, Tu, He, Wu, Deng, Ge, Chen, Zhang, Wang, Dou, Lu, Zhu, Lu, Lin, Qiao, Dai, and Wang}]{chen2025expandingperformanceboundariesopensource}
Zhe Chen, Weiyun Wang, Yue Cao, Yangzhou Liu, Zhangwei Gao, Erfei Cui, Jinguo Zhu, Shenglong Ye, Hao Tian, Zhaoyang Liu, Lixin Gu, Xuehui Wang, Qingyun Li, Yimin Ren, Zixuan Chen, Jiapeng Luo, Jiahao Wang, Tan Jiang, Bo~Wang, and 23 others. 2025{\natexlab{b}}.
\newblock \href {https://arxiv.org/abs/2412.05271} {Expanding performance boundaries of open-source multimodal models with model, data, and test-time scaling}.
\newblock \emph{Preprint}, arXiv:2412.05271.

\bibitem[{Chen et~al.(2024)Chen, Wu, Zhou, Wen, Bi, Jiang, Cao, Hu, Lai, Xiong, and Huang}]{chen2024tombench}
Zhuang Chen, Jincenzi Wu, Jinfeng Zhou, Bosi Wen, Guanqun Bi, Gongyao Jiang, Yaru Cao, Mengting Hu, Yunghwei Lai, Zexuan Xiong, and Minlie Huang. 2024.
\newblock \href {https://doi.org/10.18653/v1/2024.acl-long.847} {{T}o{MB}ench: Benchmarking theory of mind in large language models}.
\newblock In \emph{Proceedings of the 62nd Annual Meeting of the Association for Computational Linguistics (Volume 1: Long Papers)}, pages 15959--15983, Bangkok, Thailand. Association for Computational Linguistics.

\bibitem[{Cheng et~al.(2024)Cheng, Leng, Zhang, Xin, Li, Chen, Zhu, Zhang, Luo, Zhao, and Bing}]{damonlpsg2024videollama2}
Zesen Cheng, Sicong Leng, Hang Zhang, Yifei Xin, Xin Li, Guanzheng Chen, Yongxin Zhu, Wenqi Zhang, Ziyang Luo, Deli Zhao, and Lidong Bing. 2024.
\newblock \href {https://arxiv.org/abs/2406.07476} {Videollama 2: Advancing spatial-temporal modeling and audio understanding in video-llms}.
\newblock \emph{arXiv preprint arXiv:2406.07476}.

\bibitem[{Chu et~al.(2024)Chu, Xu, Yang, Wei, Wei, Guo, Leng, Lv, He, Lin, Zhou, and Zhou}]{chu2024qwen2audiotechnicalreport}
Yunfei Chu, Jin Xu, Qian Yang, Haojie Wei, Xipin Wei, Zhifang Guo, Yichong Leng, Yuanjun Lv, Jinzheng He, Junyang Lin, Chang Zhou, and Jingren Zhou. 2024.
\newblock \href {https://arxiv.org/abs/2407.10759} {Qwen2-audio technical report}.
\newblock \emph{Preprint}, arXiv:2407.10759.

\bibitem[{De~Sonneville et~al.(2002)De~Sonneville, Verschoor, Njiokiktjien, Op~het Veld, Toorenaar, and Vranken}]{desoneville2002facial}
LMJ De~Sonneville, CA~Verschoor, C~Njiokiktjien, V~Op~het Veld, N~Toorenaar, and M~Vranken. 2002.
\newblock Facial identity and facial emotions: speed, accuracy, and processing strategies in children and adults.
\newblock \emph{Journal of Clinical and experimental neuropsychology}, 24(2):200--213.

\bibitem[{Ghermi et~al.(2025)Ghermi, Wang, Kalogeiton, and Laptev}]{ghermi2025longstoryshortstorylevel}
Ridouane Ghermi, Xi~Wang, Vicky Kalogeiton, and Ivan Laptev. 2025.
\newblock \href {https://arxiv.org/abs/2406.10221} {Long story short: Story-level video understanding from 20k short films}.
\newblock \emph{Preprint}, arXiv:2406.10221.

\bibitem[{Guo et~al.(2023)Guo, Li, and Haf}]{guo-etal-2023-desiq}
Xiao-Yu Guo, Yuan-Fang Li, and Reza Haf. 2023.
\newblock \href {https://doi.org/10.18653/v1/2023.emnlp-main.191} {{D}e{SIQ}: Towards an unbiased, challenging benchmark for social intelligence understanding}.
\newblock In \emph{Proceedings of the 2023 Conference on Empirical Methods in Natural Language Processing}, pages 3169--3180, Singapore. Association for Computational Linguistics.

\bibitem[{Jin et~al.(2024)Jin, Wu, Cao, Xiang, Kuo, Hu, Ullman, Torralba, Tenenbaum, and Shu}]{jin2024mmtom}
Chuanyang Jin, Yutong Wu, Jing Cao, Jiannan Xiang, Yen-Ling Kuo, Zhiting Hu, Tomer Ullman, Antonio Torralba, Joshua~B Tenenbaum, and Tianmin Shu. 2024.
\newblock Mmtom-qa: Multimodal theory of mind question answering.
\newblock \emph{arXiv preprint arXiv:2401.08743}.

\bibitem[{KimiTeam et~al.(2025)KimiTeam, Ding, Ju, Leng, Liu, Liu, Shang, Shen, Song, Tan, Tang, Wang, Wei, Xin, Xu, Yu, Zhang, Zhou, Charles, Chen, Chen, Du, He, Hu, Lai, Li, Liu, Sun, Wang, Wang, Wu, Wu, Yang, Yang, Yang, Yang, Yin, Yuan, Zhang, and Zhou}]{kimiteam2025kimiaudiotechnicalreport}
KimiTeam, Ding Ding, Zeqian Ju, Yichong Leng, Songxiang Liu, Tong Liu, Zeyu Shang, Kai Shen, Wei Song, Xu~Tan, Heyi Tang, Zhengtao Wang, Chu Wei, Yifei Xin, Xinran Xu, Jianwei Yu, Yutao Zhang, Xinyu Zhou, Y.~Charles, and 21 others. 2025.
\newblock \href {https://arxiv.org/abs/2504.18425} {Kimi-audio technical report}.
\newblock \emph{Preprint}, arXiv:2504.18425.

\bibitem[{Le et~al.(2019)Le, Boureau, and Nickel}]{le2019revisiting}
Matthew Le, Y-Lan Boureau, and Maximilian Nickel. 2019.
\newblock Revisiting the evaluation of theory of mind through question answering.
\newblock In \emph{Proceedings of the 2019 Conference on Empirical Methods in Natural Language Processing and the 9th International Joint Conference on Natural Language Processing (EMNLP-IJCNLP)}, pages 5872--5877.

\bibitem[{Ma et~al.(2023)Ma, Sansom, Peng, and Chai}]{ma-etal-2023-towards-holistic}
Ziqiao Ma, Jacob Sansom, Run Peng, and Joyce Chai. 2023.
\newblock \href {https://doi.org/10.18653/v1/2023.findings-emnlp.72} {Towards a holistic landscape of situated theory of mind in large language models}.
\newblock In \emph{Findings of the Association for Computational Linguistics: EMNLP 2023}, pages 1011--1031, Singapore. Association for Computational Linguistics.

\bibitem[{Mathur et~al.(2025)Mathur, Qian, Liang, and Morency}]{mathur2025socialgenomegroundedsocial}
Leena Mathur, Marian Qian, Paul~Pu Liang, and Louis-Philippe Morency. 2025.
\newblock \href {https://arxiv.org/abs/2502.15109} {Social genome: Grounded social reasoning abilities of multimodal models}.
\newblock \emph{Preprint}, arXiv:2502.15109.

\bibitem[{Oguntola et~al.(2021)Oguntola, Hughes, and Sycara}]{oguntola2021deepinterpretablemodelstheory}
Ini Oguntola, Dana Hughes, and Katia Sycara. 2021.
\newblock \href {https://arxiv.org/abs/2104.02938} {Deep interpretable models of theory of mind}.
\newblock \emph{Preprint}, arXiv:2104.02938.

\bibitem[{Premack and Woodruff(1978)}]{premack1978does}
David Premack and Guy Woodruff. 1978.
\newblock Does the chimpanzee have a theory of mind?
\newblock \emph{Behavioral and brain sciences}, 1(4):515--526.

\bibitem[{Sabour et~al.(2024)Sabour, Liu, Zhang, Liu, Zhou, Sunaryo, Lee, Mihalcea, and Huang}]{sabour2024emobench}
Sahand Sabour, Siyang Liu, Zheyuan Zhang, June Liu, Jinfeng Zhou, Alvionna Sunaryo, Tatia Lee, Rada Mihalcea, and Minlie Huang. 2024.
\newblock \href {https://doi.org/10.18653/v1/2024.acl-long.326} {{E}mo{B}ench: Evaluating the emotional intelligence of large language models}.
\newblock In \emph{Proceedings of the 62nd Annual Meeting of the Association for Computational Linguistics (Volume 1: Long Papers)}, pages 5986--6004, Bangkok, Thailand. Association for Computational Linguistics.

\bibitem[{Sap et~al.(2019)Sap, Rashkin, Chen, Le~Bras, and Choi}]{sap2019socialiqa}
Maarten Sap, Hannah Rashkin, Derek Chen, Ronan Le~Bras, and Yejin Choi. 2019.
\newblock \href {https://doi.org/10.18653/v1/D19-1454} {Social {IQ}a: Commonsense reasoning about social interactions}.
\newblock In \emph{Proceedings of the 2019 Conference on Empirical Methods in Natural Language Processing and the 9th International Joint Conference on Natural Language Processing (EMNLP-IJCNLP)}, pages 4463--4473, Hong Kong, China. Association for Computational Linguistics.

\bibitem[{Wilf et~al.(2023)Wilf, Mathur, Mathew, Ko, Kebe, Liang, and Morency}]{siq2}
Alex Wilf, Leena Mathur, Sheryl Mathew, Claire Ko, Youssouf Kebe, Paul~Pu Liang, and Louis-Philippe Morency. 2023.
\newblock Social-iq 2.0 challenge: Benchmarking multimodal social understanding.
\newblock \url{https://github.com/abwilf/Social-IQ-2.0-Challenge}.

\bibitem[{Williams et~al.(2022)Williams, Fiore, and Jentsch}]{williams2022supporting}
Jessica Williams, Stephen~M Fiore, and Florian Jentsch. 2022.
\newblock Supporting artificial social intelligence with theory of mind.
\newblock \emph{Frontiers in artificial intelligence}, 5:750763.

\bibitem[{Wolf et~al.(2020)Wolf, Debut, Sanh, Chaumond, Delangue, Moi, Cistac, Rault, Louf, Funtowicz, Davison, Shleifer, von Platen, Ma, Jernite, Plu, Xu, Scao, Gugger, Drame, Lhoest, and Rush}]{wolf2020huggingfacestransformersstateoftheartnatural}
Thomas Wolf, Lysandre Debut, Victor Sanh, Julien Chaumond, Clement Delangue, Anthony Moi, Pierric Cistac, Tim Rault, Rémi Louf, Morgan Funtowicz, Joe Davison, Sam Shleifer, Patrick von Platen, Clara Ma, Yacine Jernite, Julien Plu, Canwen Xu, Teven~Le Scao, Sylvain Gugger, and 3 others. 2020.
\newblock \href {https://arxiv.org/abs/1910.03771} {Huggingface's transformers: State-of-the-art natural language processing}.
\newblock \emph{Preprint}, arXiv:1910.03771.

\bibitem[{Wu et~al.(2023)Wu, He, Jia, Mihalcea, Chen, and Deng}]{he2023hi}
Yufan Wu, Yinghui He, Yilin Jia, Rada Mihalcea, Yulong Chen, and Naihao Deng. 2023.
\newblock \href {https://doi.org/10.18653/v1/2023.findings-emnlp.717} {Hi-{T}o{M}: A benchmark for evaluating higher-order theory of mind reasoning in large language models}.
\newblock In \emph{Findings of the Association for Computational Linguistics: EMNLP 2023}, pages 10691--10706, Singapore. Association for Computational Linguistics.

\bibitem[{Xu et~al.(2025)Xu, Guo, He, Hu, He, Bai, Chen, Wang, Fan, Dang et~al.}]{Qwen2.5-Omni}
Jin Xu, Zhifang Guo, Jinzheng He, Hangrui Hu, Ting He, Shuai Bai, Keqin Chen, Jialin Wang, Yang Fan, Kai Dang, and 1 others. 2025.
\newblock Qwen2. 5-omni technical report.
\newblock \emph{arXiv preprint arXiv:2503.20215}.

\bibitem[{Yao et~al.(2024)Yao, Yu, Zhang, Wang, Cui, Zhu, Cai, Li, Zhao, He et~al.}]{yao2024minicpm}
Yuan Yao, Tianyu Yu, Ao~Zhang, Chongyi Wang, Junbo Cui, Hongji Zhu, Tianchi Cai, Haoyu Li, Weilin Zhao, Zhihui He, and 1 others. 2024.
\newblock Minicpm-v: A gpt-4v level mllm on your phone.
\newblock \emph{arXiv preprint arXiv:2408.01800}.

\bibitem[{Ye and Kovashka(2021)}]{shortcuts}
Keren Ye and Adriana Kovashka. 2021.
\newblock A case study of the shortcut effects in visual commonsense reasoning.
\newblock In \emph{Proceedings of the AAAI conference on artificial intelligence}, volume~35, pages 3181--3189.

\bibitem[{Zhang et~al.(2024{\natexlab{a}})Zhang, Zhang, Li, Zeng, Yang, Zhang, Wang, Tan, Li, and Liu}]{zhang2024longcontexttransferlanguage}
Peiyuan Zhang, Kaichen Zhang, Bo~Li, Guangtao Zeng, Jingkang Yang, Yuanhan Zhang, Ziyue Wang, Haoran Tan, Chunyuan Li, and Ziwei Liu. 2024{\natexlab{a}}.
\newblock \href {https://arxiv.org/abs/2406.16852} {Long context transfer from language to vision}.
\newblock \emph{Preprint}, arXiv:2406.16852.

\bibitem[{Zhang et~al.(2024{\natexlab{b}})Zhang, Wu, Li, Li, Ma, Liu, and Li}]{zhang2024videoinstructiontuningsynthetic}
Yuanhan Zhang, Jinming Wu, Wei Li, Bo~Li, Zejun Ma, Ziwei Liu, and Chunyuan Li. 2024{\natexlab{b}}.
\newblock \href {https://arxiv.org/abs/2410.02713} {Video instruction tuning with synthetic data}.
\newblock \emph{Preprint}, arXiv:2410.02713.

\bibitem[{Zhou et~al.(2023)Zhou, Zhu, Mathur, Zhang, Yu, Qi, Morency, Bisk, Fried, Neubig et~al.}]{zhou2023sotopia}
Xuhui Zhou, Hao Zhu, Leena Mathur, Ruohong Zhang, Haofei Yu, Zhengyang Qi, Louis-Philippe Morency, Yonatan Bisk, Daniel Fried, Graham Neubig, and 1 others. 2023.
\newblock Sotopia: Interactive evaluation for social intelligence in language agents.
\newblock \emph{arXiv preprint arXiv:2310.11667}.

\end{thebibliography}

\appendix

\section{Appendix}
\label{sec:appendix}

\subsection{Samples from \textsc{MoMentS} Across ToM Abilities}
\label{sec:moments_examples}

\begin{figure*}
  \centering
  \includegraphics[trim={0cm 0cm 0cm 0cm},clip,width=0.9\linewidth]{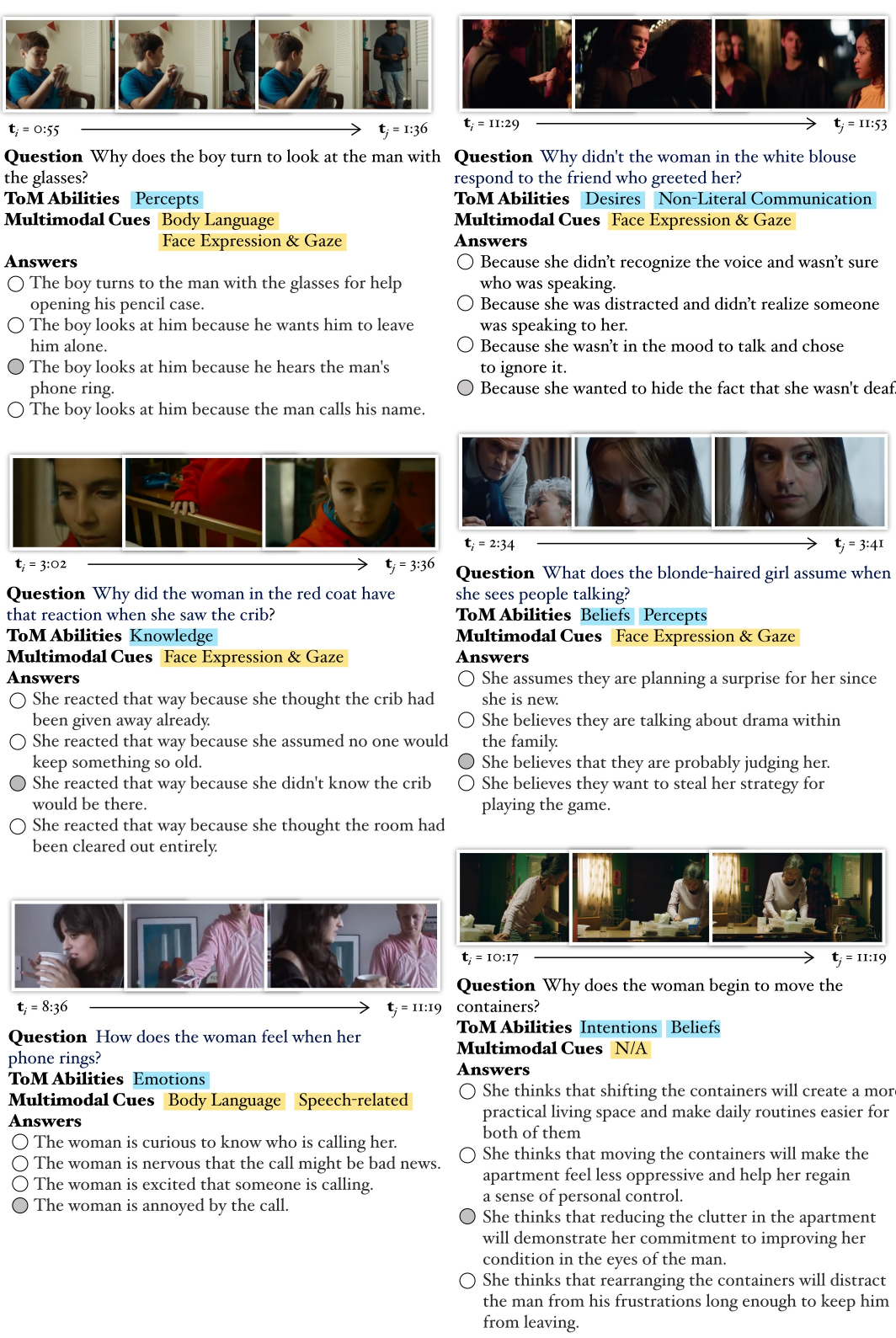}
  \caption{Samples from \textsc{MoMentS} Representing Each ToM Abilities.}
  \label{fig:examples}
\end{figure*}

Figure \ref{fig:examples} presents representative samples of \textsc{MoMentS} questions covering different ToM abilities. Each example includes the question, the full answer set (one correct option and three distractors), the targeted ToM abilities, and any multimodal cues identified by annotators as relevant for answering the question.


\subsection{Pilot Annotations}
\label{sec:pilot}

We conducted two pilot annotation phases prior to the main annotation batch to identify challenges and refine our pipeline.

\paragraph{First Pilot Annotation}  
We recruited annotators through Prolific, selecting participants who were native English speakers with a university degree. Each annotator was asked to create both questions and distractors covering all seven ToM abilities. This pilot produced 268 question–answer sets. From analyzing submissions from this annotation batch, we identified the following issues:

\begin{itemize}
\item Many questions were low quality, some had grammatical issues, others focused on plot rather than ToM.
\item Models achieved over 50\% accuracy without context, pointing to biases in the distractor sets (see Table \ref{tab:biases}).
\item Annotators often mislabeled the ToM ability, indicating limited understanding of the categories.
\end{itemize}

We traced these problems to the following causes:
\begin{itemize}
\item Time constraints imposed by Prolific's system created pressure that negatively impacted annotation quality. 
\item Prolific communication channels made direct communication with annotators difficult, as they did not communicate their questions effectively. 
\item Tasking annotators with all seven categories was overwhelming, leading to overall misclassification.
\item Most effort was spent on writing questions, resulting in weaker distractors.
\item Models could exploit biases in seemingly good distractors, without needing any context to answer.
\end{itemize}

\paragraph{Second Pilot Annotation}  
To address these issues, we made the following changes:

\begin{itemize}
\item We directly hired seven undergraduate students from psychology and social sciences and used group messaging for better communication.
\item Each annotator was assigned only 2–3 ToM abilities to help them specialize.
\item Annotation was split into two phases: creating questions in the first week and distractors in the second. This was done to help annotators concentrate their efforts on writing high-quality questions first, then shift their focus to creating high-quality distractors.
\item A custom annotation platform with an LLM was introduced to automatically flag biased distractors (see Section \ref{sec:copilot}).
\item Annotators were encouraged to spread their work throughout the week to reduce low-quality submissions due to pressure in last-minute submissions.
\item We provided weekly reviews and feedback to improve consistency and quality.
\end{itemize}

This second pilot resulted in 350 high-quality questions. Most of the design choices from this phase were carried over to the main annotation batch.


\subsection{Prompt For Video Filtering}
\label{sec:video_filtering}

\begin{lstlisting}[basicstyle=\ttfamily\small, breaklines=true, frame=single]
You are a film critic and psychologist with expertise in Theory of Mind (ToM) as described by the ATOMS taxonomy. Your task is to analyze the movie synopsis and captions below to determine how likely it is that the movie includes themes or questions related to Theory of Mind.

Theory of Mind involves understanding and attributing mental states to oneself and others. Consider the following key components:
1. Knowledge: Recognizing that characters hold organized information and mental representations that shape their understanding.
2. Emotions: Identifying complex emotional responses, including mixed or evolving emotions.
3. Desires: Understanding that characters may have varied and sometimes conflicting desires driving their actions.
4. Beliefs: Discerning true versus false beliefs and recognizing higher-order beliefs (beliefs about others' beliefs).
5. Intentions: Inferring characters' goals and the reasoning behind their actions.
6. Percepts: Noting how characters perceive their world differently based on their sensory experiences.
7. Non-literal Communication: Interpreting subtleties such as sarcasm, humor, or metaphors that imply meanings beyond the literal words.

Using this framework, please analyze the following content:

Movie Synopsis: {synopsis}
Movie Captions: {caption}

Based on your analysis, provide a probability (as an integer percentage between 0 and 100) indicating how likely it is that this movie involves Theory of Mind related questions or themes. Your answer should be only the integer value with no additional commentary. choose the best number that seems appropriate based on the data.
\end{lstlisting}

\subsection{Evaluation on ASR quality}
\label{sec:asr}

In this subsection, we describe our audio processing pipeline, present, and report its ASR performance on a subset of human-annotated transcripts.
\begin{table*}[]
\centering
\resizebox{1.4\columnwidth}{!}{%
\begin{tabular}{lllllllllll}
\toprule
                              & \multicolumn{4}{c}{English}   &  &  & \multicolumn{4}{c}{Non-English} \\
\midrule
Model                     & \multicolumn{2}{c}{$[t_0,t_j]$} & \multicolumn{2}{c}{$[t_i,t_j]$} &  &  & \multicolumn{2}{c}{$[t_0,t_j]$} & \multicolumn{2}{c}{$[t_i,t_j]$} \\
\midrule
                              & $T$   & $VT$  & $T$   & $VT$  &  &  & $T$    & $VT$   & $T$   & $VT$  \\
\midrule
\ding{108} LLaVA-Video-7B     & 47.59 & 49.11 & 46.53 & 51.39 &  &  & 43.90  & 50.68  & 40.38 & 55.28 \\
\ding{108} InternVL2.5 8B     & 46.33 & 46.84 & 45.06 & 50.84 &  &  & 44.17  & 45.53  & 47.43 & 56.91 \\
\ding{108} LongVA-7B-DPO      & 41.37 & 43.85 & 41.11 & 44.71 &  &  & 38.75  & 46.34  & 43.36 & 43.36 \\
\ding{108} Qwen2.5 VL 7B      & 40.71 & 38.23 & 37.92 & 44.15 &  &  & 44.72  & 37.13  & 40.92 & 45.26 \\
\ding{108} LLaVA-Video-72B    & 62.99 & 66.08 & 62.58 & 67.59 &  &  & 64.23  & 65.31  & 59.35 & 68.02 \\
\ding{108} InternVL2.5 78B    & 53.72 & 61.11 & 52.35 & 60.46 &  &  & 51.22  & 60.98  & 51.49 & 66.94 \\
\midrule
\ding{108} Average            & 48.78 & 50.87 & 47.59 & 53.19 &  &  & 47.83  & 50.99  & 47.15 & 55.96 \\
\midrule
                              & $T$   & $VT$  & $T$   & $VT$  &  &  & $T$    & $VT$   & $T$   & $VT$  \\
\midrule
\ding{115} Qwen2.5-Omni-7B    & 46.38 & 52.71 & 45.57 & 54.99 &  &  & 49.32  & 57.18  & 47.15 & 62.60 \\
\ding{115} VideoLLaMA2-7B-AV  & 36.76 & 40.66 & 38.58 & 42.99 &  &  & 42.28  & 42.55  & 37.40 & 43.90 \\
\ding{115} MiniCPM-o 2.6 (8B) & 47.70 &       & 46.99 & 49.37 &  &  & 48.51  &        & 47.43 & 54.47 \\
\midrule
\ding{115} Average            & 43.61 & 46.68 & 43.71 & 49.11 &  &  & 46.70  & 49.86  & 43.99 & 53.66 \\
\midrule
                              & $A$   & $VA$  & $A$   & $VA$  &  &  & $A$    & $VA$   & $A$   & $VA$  \\
\midrule
\ding{115} Qwen2.5-Omni-7B    & 44.30 & 52.15 & 48.91 & 54.28 &  &  & 44.99  & 55.56  & 46.07 & 62.60 \\
\ding{115} VideoLLaMA2-7B-AV  & 33.77 & 41.92 & 33.82 & 42.99 &  &  & 36.59  & 44.44  & 36.31 & 46.88 \\
\ding{115} MiniCPM-o 2.6 (8B) & 39.54 & **    & 40.00 & 48.10 &  &  & 39.02  & **     & 37.67 & 49.05 \\
\midrule
\ding{115} Average            & 39.21 & 47.04 & 40.91 & 48.46 &  &  & 40.20  & 50.00  & 40.02 & 52.85 \\
\midrule
                          & \multicolumn{2}{l}{$A$}         & \multicolumn{2}{l}{$A$}         &  &  & \multicolumn{2}{l}{$A$}         & \multicolumn{2}{l}{$A$}         \\
\midrule
\ding{116} Kimi-Audio-7B  & \multicolumn{2}{l}{31.95}       & \multicolumn{2}{l}{48.41}       &  &  & \multicolumn{2}{l}{30.08}       & \multicolumn{2}{l}{49.59}       \\
\ding{116} Qwen2-Audio-7B & \multicolumn{2}{l}{35.54}       & \multicolumn{2}{l}{35.65}       &  &  & \multicolumn{2}{l}{29.00}       & \multicolumn{2}{l}{34.42}       \\
\midrule
\ding{116} Average        & \multicolumn{2}{l}{33.75}       & \multicolumn{2}{l}{42.03}       &  &  & \multicolumn{2}{l}{29.54}       & \multicolumn{2}{l}{42.01} \\
\bottomrule
\end{tabular}%
}
\caption{Global accuracy of different Video LLMs (\ding{108}), Audiovisual LLMs (\ding{115}), and Speech LLMs (\ding{116}) across English and non-English videos.}
\label{tab:multilingual_global_acc}
\end{table*}

\paragraph{ASR Pipeline} We use WhisperX \cite{whisperx} to transcribe the short films. Its multilingual capabilities make it suitable for both English and non-English videos in our dataset. For speaker diarization, we employ PyAnnote \cite{pyannote}.

\paragraph{ASR Quality Evaluation}
We evaluate the ASR pipeline using different base Whisper models on a subset of 50 human-transcribed videos, reporting global \textbf{Word-Error Rate (WER)} and \textbf{Diarization Error Rate (DER)}. 
For global WER we concatenate each file's reference and ASR transcripts lower-casing and punctuation removal and computing 
$\text{WER} = (S + I + D) / N$, 
where $S$, $I$, and $D$ are the numbers of substituted, inserted, and deleted words, and $N$ is the total number of reference words. For DER, we evaluate only within spans where the reference marks speech. The score is 
$\text{DER} = (T_{\text{missed}} + T_{\text{confusion}}) / T_{\text{ref}}$, 
where $T_{\text{missed}}$ is reference speech with no ASR coverage, $T_{\text{confusion}}$ is overlapped speech attributed to the wrong mapped speaker, and $T_{\text{ref}}$ is the total duration of speech in the reference annotation. We report these in Table \ref{tab:wer}, while Whisper large-v3 scores the lowest average global WER, in practice we notice that it failed to transcribe some of the videos. This does not happen with \textit{large-v2}, whose DER is the lowest; because of this, we opted for the latter as the chosen model for transcribing audio for the Video LLMs.

\begin{table}[]
\centering
\resizebox{0.65\columnwidth}{!}{%
\begin{tabular}{lll}
\toprule
 & global-WER & DER \\
 \midrule
base & 36.2 & 41.2 \\
large-v2 & 20.6 & 36.3 \\
large-v3 & 16.6 & 40.9 \\
\bottomrule
\end{tabular}%
}
\caption{Comparison of average WER and DER across the three evaluated models.}
\label{tab:wer}
\end{table}

\subsection{Ablation on number of frames}
\label{sec:frame_ablation}

Increasing the number of video frames increases computational cost, as most Video LLMs embed frame patches significantly extending the context length processed by the language model. To assess the tradeoff between context length and performance, we evaluate three models on 1,500 randomly selected \textsc{MoMentS} entries using 64 and 96 frames.

As shown in Table \ref{tab:frame_ablations}, increasing the number of frames does not lead to consistent improvements. In several cases, performance actually drops, likely due to redundancy or context saturation. Based on these results, we use 64 frames for all main Video-LLMs evaluations in the paper.

\begin{table*}[]
\centering
\resizebox{0.7\linewidth}{!}{%
\begin{tabular}{lllllll}
\toprule
 & \multicolumn{3}{c}{$[t_0,t_j]$} & \multicolumn{3}{c}{$[t_i,t_j]$} \\
Model & $T$ & $VT$-64 & $VT$-96 & $T$ & $VT$-64 & $VT$-96 \\
\midrule
LLaVA-Video-7B & 46.33 & 47.7 (\textbf{+1.4}) & 47.3 (+1.0) & 44.45 & 50.7 (\textbf{+6.3}) & 49.7 (+5.3) \\
LongVA-7B-DPO & 40.94 & 45.5 (\textbf{+4.5}) & 42.6 (+1.6) & 41.19 & 42.9 (+1.8) & 44.6 (\textbf{+3.4}) \\
InternVL2.5 8B & 45.58 & 45.6 (+0.1) & 48.2 (\textbf{+2.6}) & 44.45 & 51.7 (\textbf{+7.3}) & 50.4 (+6.0) \\
\bottomrule
\end{tabular}%
}
\caption{Global accuracy on a subset of 1,500 \textsc{MoMentS} samples using only transcripts ($T$), and transcripts plus 64 or 96 frames ($VT$-64 and $VT$-96). Results are reported for both the Full ($[t_0, t_j]$) and Focused ($[t_i, t_j]$) Context Windows. We mark in bold the highest increase over $T$ between 96 and 64 frames.}
\label{tab:frame_ablations}
\end{table*}

\subsection{Performance comparison in English and Non-English videos.}

As noted in Appendix \ref{sec:data_stats}, \textsc{MoMentS} includes a subset of non-English videos. Table \ref{tab:multilingual_global_acc} compares performance on English-only and non-English clips. We find no substantial drops in accuracy for non-English videos; in fact, models with visual inputs often perform better in this setting. A possible explanation for this is that most of the non-English videos include subtitles in the frames, which may support the temporal grounding of dialogues.

\subsection{Dataset Statistics and Annotation Cost}
\label{sec:data_stats}

The main annotation batch involved 16 participants: 12 undergraduate students in psychology and social sciences, two computer scientists, and two clinical psychologists. 12 of them were female and 4 male, all of them between 20 and 30 years old. Twelve of the annotators were from Canada, and the remaining were from Mexico. All participants were explained the purpose of their annotations in an onboarding session. The annotation process received approval from the MBZUAI Ethics Review Board.

\paragraph{Annotation Cost}
Annotators were compensated at a rate of 17 CAD per hour through UpWork. To encourage steady progress, a weekly bonus of 10 USD was provided to those who completed at least half of their assignments by midweek. An additional performance-based bonus of 150 USD was awarded to annotators who produced the highest-quality annotations. The total cost of the \textsc{MoMentS} main annotation effort amounted to 8,745 USD.

\paragraph{Dataset Statistics}

\begin{table}[]
\centering
\resizebox{0.5\columnwidth}{!}{%
\begin{tabular}{lr}
        \toprule
        \textbf{Language} & \textbf{Number of Videos} \\
        \midrule
        English   & 144 \\
        Russian   & 6   \\
        Spanish   & 5   \\
        French    & 3   \\
        Persian   & 3   \\
        Italian   & 1   \\
        Arabic    & 1   \\
        Swedish   & 1   \\
        Korean    & 1   \\
        Danish    & 1   \\
        Hindi     & 1   \\
        Japanese  & 1   \\
        \midrule
        \textbf{Total} & \textbf{168} \\
        \bottomrule
    \end{tabular}
}
\caption{Number of videos per language. }
\label{tab:languages}
\end{table}

\textsc{MoMentS} contains 2,335 questions across 168 short films, the majority of which are in English (144). We also include a subset of 24 films in other languages. Table \ref{tab:languages} reports the number of videos per language.

\begin{figure*}[t]                
  \centering
   \begin{subcaptionbox}{Distribution of lengths for correct answers and distractors.\label{fig:answer_distributions}}
        [0.3\linewidth]
        {\includegraphics[width=\linewidth]{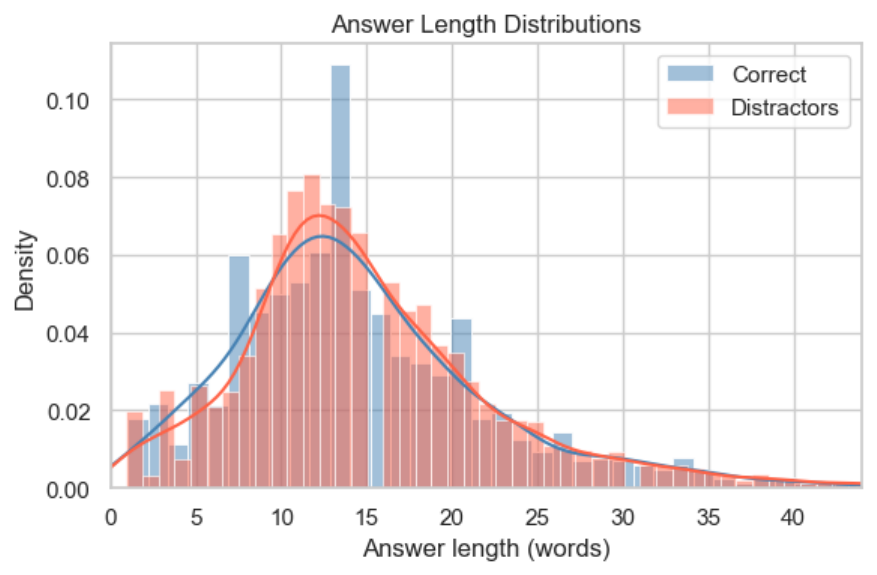}}
  \end{subcaptionbox}
  \hfill
  \begin{subcaptionbox}{Length distribution of the Focused Context Windows.\label{fig:t0t_distribution}}    
        [0.3\linewidth]
        {\includegraphics[width=\linewidth]{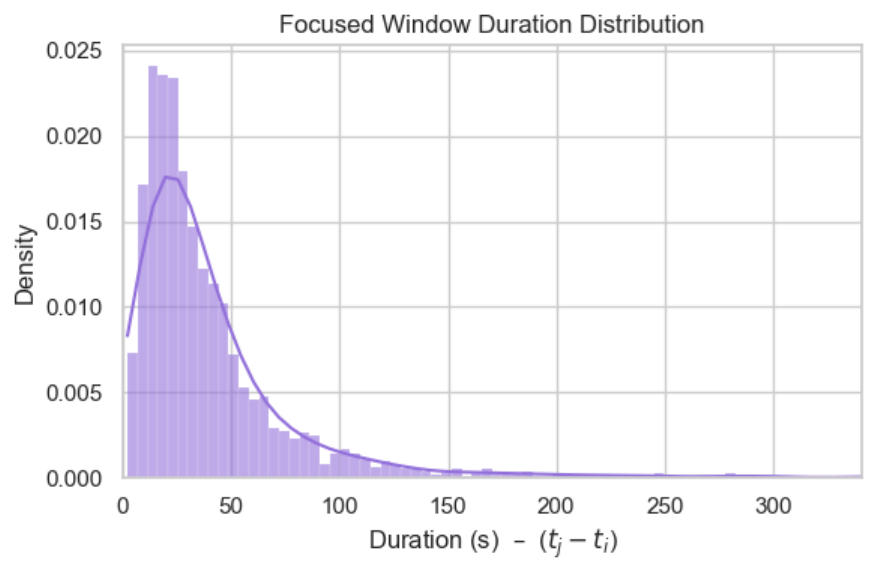}}
  \end{subcaptionbox}
  \hfill
  \begin{subcaptionbox}{Length distribution of the Full Context Windows.\label{fig:0t_distribution}}
        [0.3\linewidth]                 
        {\includegraphics[width=\linewidth]{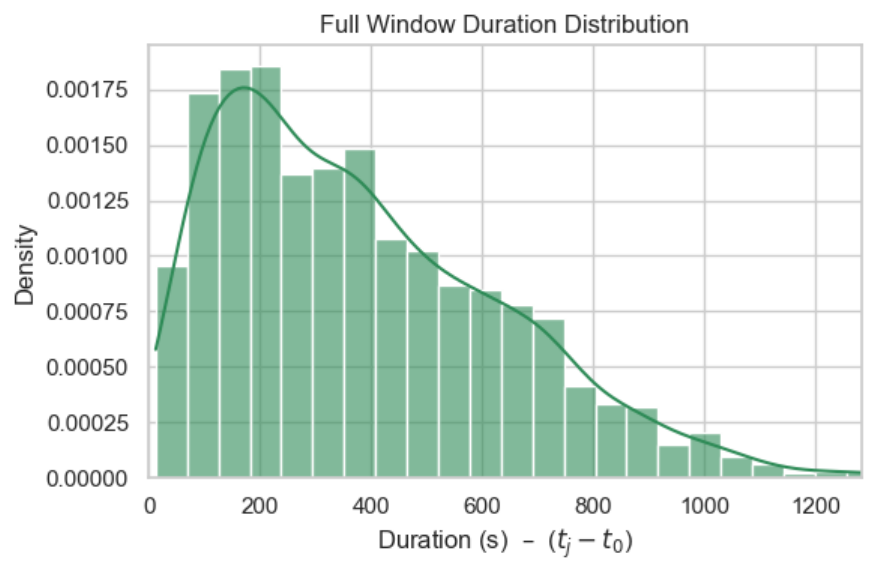}}
  \end{subcaptionbox}                              
 
  \caption{Histograms of different the statistics reported in Table \ref{tab:statistics}.}
  \label{fig:distributions}
\end{figure*}

In Table \ref{tab:statistics}, we report the average question length, average answer lengths, and durations of the full and Focused Context Windows. We also display the distributions of lengths for answers, Focused, and Full context windows in Figures \ref{fig:answer_distributions}, Figure \ref{fig:t0t_distribution}, and Figure \ref{fig:0t_distribution}, respectively.

\subsection{Copyright and License}
\label{sec:license}

We release \textsc{MoMentS} annotations under a \textbf{CC BY-NC-SA 4.0} license (Attribution-NonCommercial-ShareAlike 4.0 International), intended only for academic research purposes.

Following \citet{ghermi2025longstoryshortstorylevel} and \citet{siq2}, we do not distribute the video content directly. We provide URLs linking to the original videos on YouTube, complying with YouTube's Terms of Service (\url{https://www.youtube.com/static?template=terms}).

\subsection{Ethical Considerations}

\paragraph{Representation and Bias} 
Most of \textsc{MoMentS} videos are in English and reflect Western cultural norms. Additionally, annotators were from Canada and Mexico, which may influence interpretations of emotions, intentions, or non-literal communication. 

\paragraph{Potential Misuse}
\textsc{MoMentS} is designed to evaluate models’ ability to infer mental states in socially grounded scenarios to foster progress in socially intelligent AI. However, ToM capabilities could also be misused to simulate deceptive, manipulative, or persuasive behavior in artificial agents. To mitigate this risk, we license the dataset for academic research only under a CC BY-NC-SA 4.0 license, and we strictly stand against any use in applications that exploit it for unethical purposes.

\begin{table}[]
\centering
\resizebox{0.5\columnwidth}{!}{%
\begin{tabular}{l r}
\toprule
\textbf{ToM Ability} & \textbf{\# Questions}\\
\midrule
Emotions                  & 599  \\
Beliefs                   & 379  \\
Desires                   & 386  \\
Intentions                & 1026 \\
Percepts                  & 316  \\
Knowledge                 & 329  \\
NLC                       & 222  \\
\bottomrule
\end{tabular}
}
\caption{Number of questions associated with each ToM ability.}
\label{tab:abilities}
\end{table}

\paragraph{Personally Identifying Information or Offensive Content}

Questions and answer sets do not contain personally identifying information as they use descriptors to refer to the characters. Since questions ask about character's mental states, they do not contain offensive content.

\newpage

\subsection{Guidelines for Question and Distractor Annotation}
\label{sec:guidelines}

The following pages contain the annotation guidelines provided to annotators during the first annotation batch. Separate documents were provided for the question creation and distractor creation stages to reflect the specific goals and challenges of each.



\begin{figure*}[t]
\centering
\includegraphics[width=0.48\textwidth]{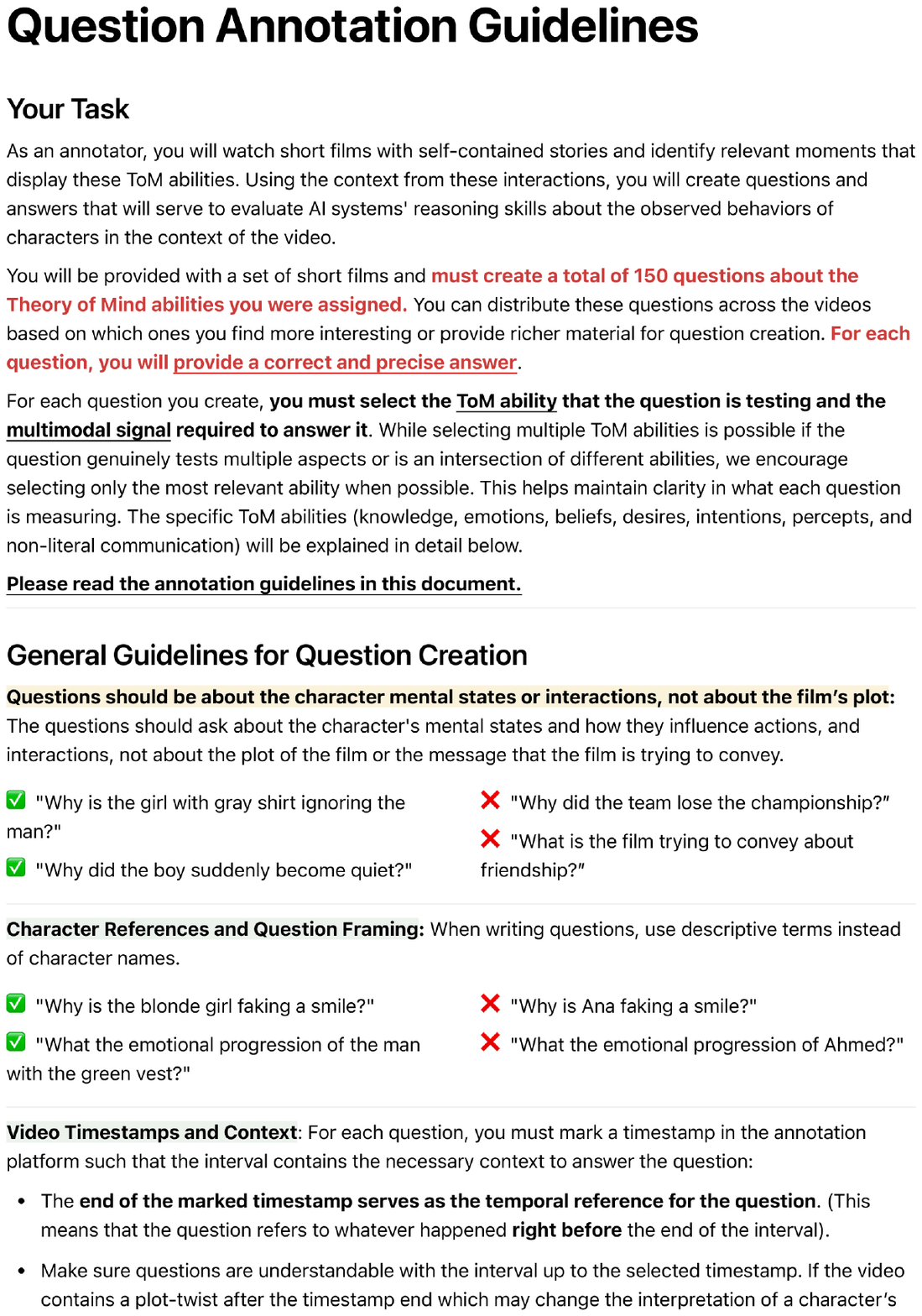}
\includegraphics[width=0.48\textwidth]{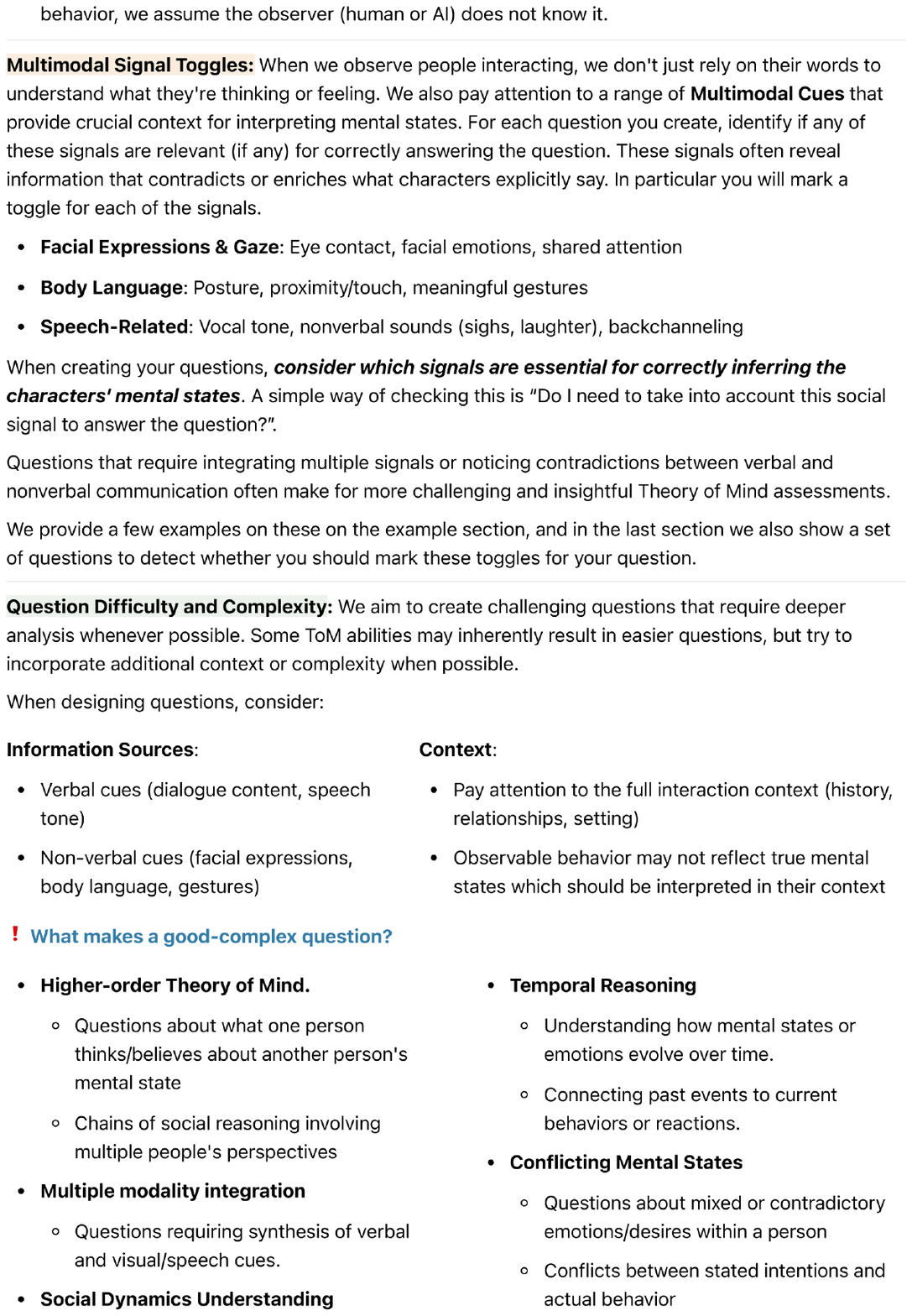}\\[0.5em]
\rule{\textwidth}{0.3pt}\\
\includegraphics[width=0.48\textwidth]{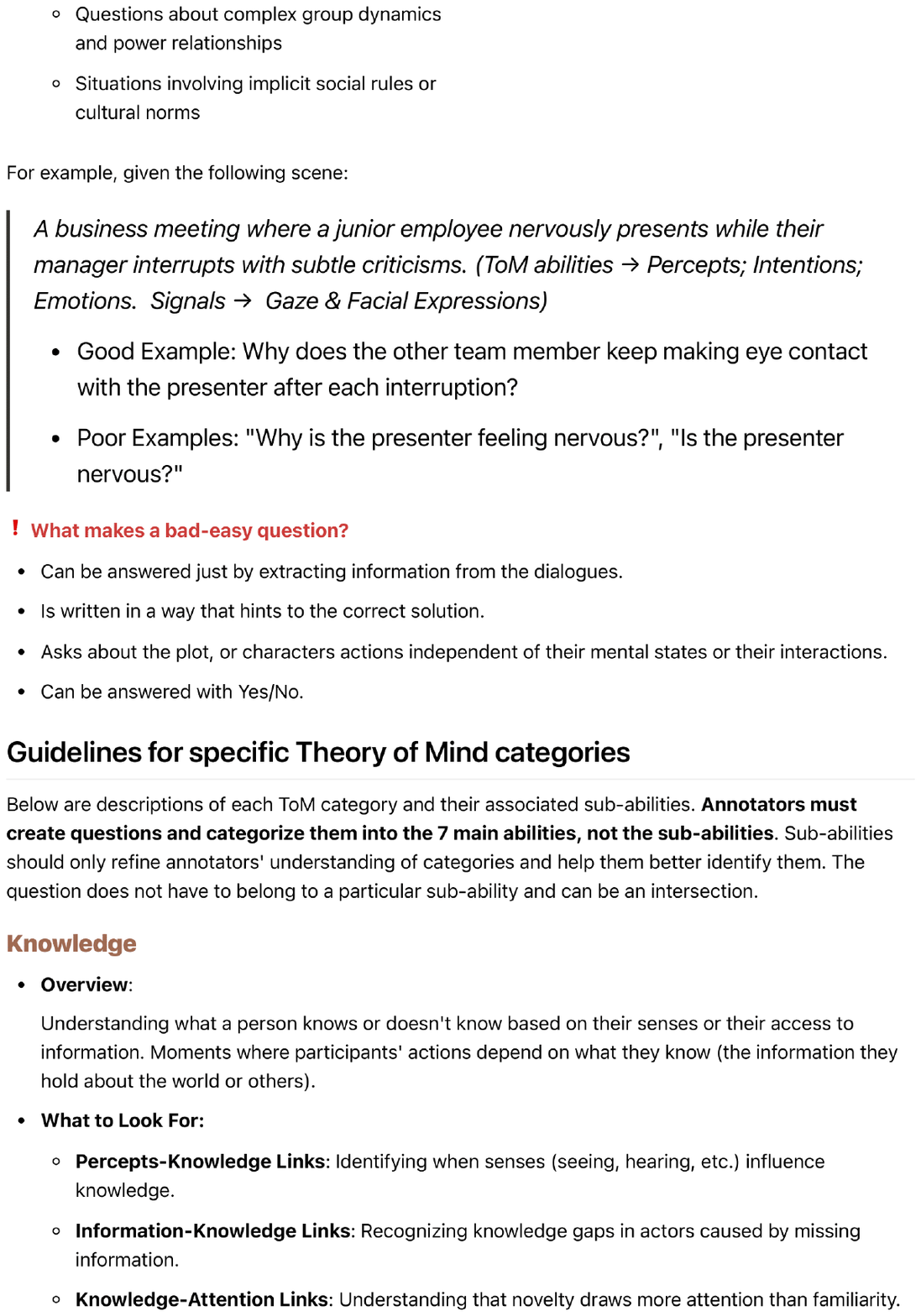}
\includegraphics[width=0.48\textwidth]{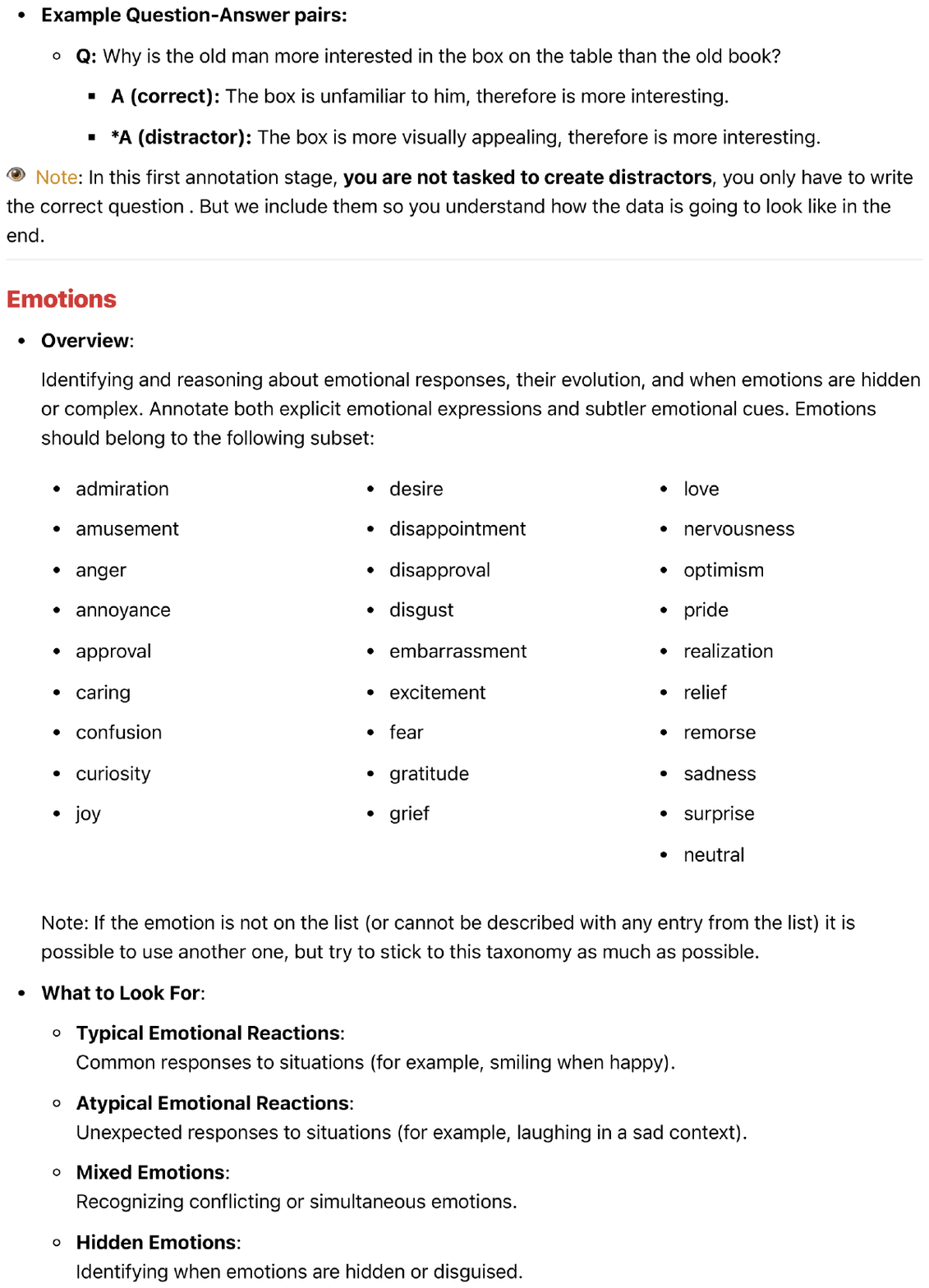}
\end{figure*}

\begin{figure*}[t]
\centering
\includegraphics[width=0.48\textwidth]{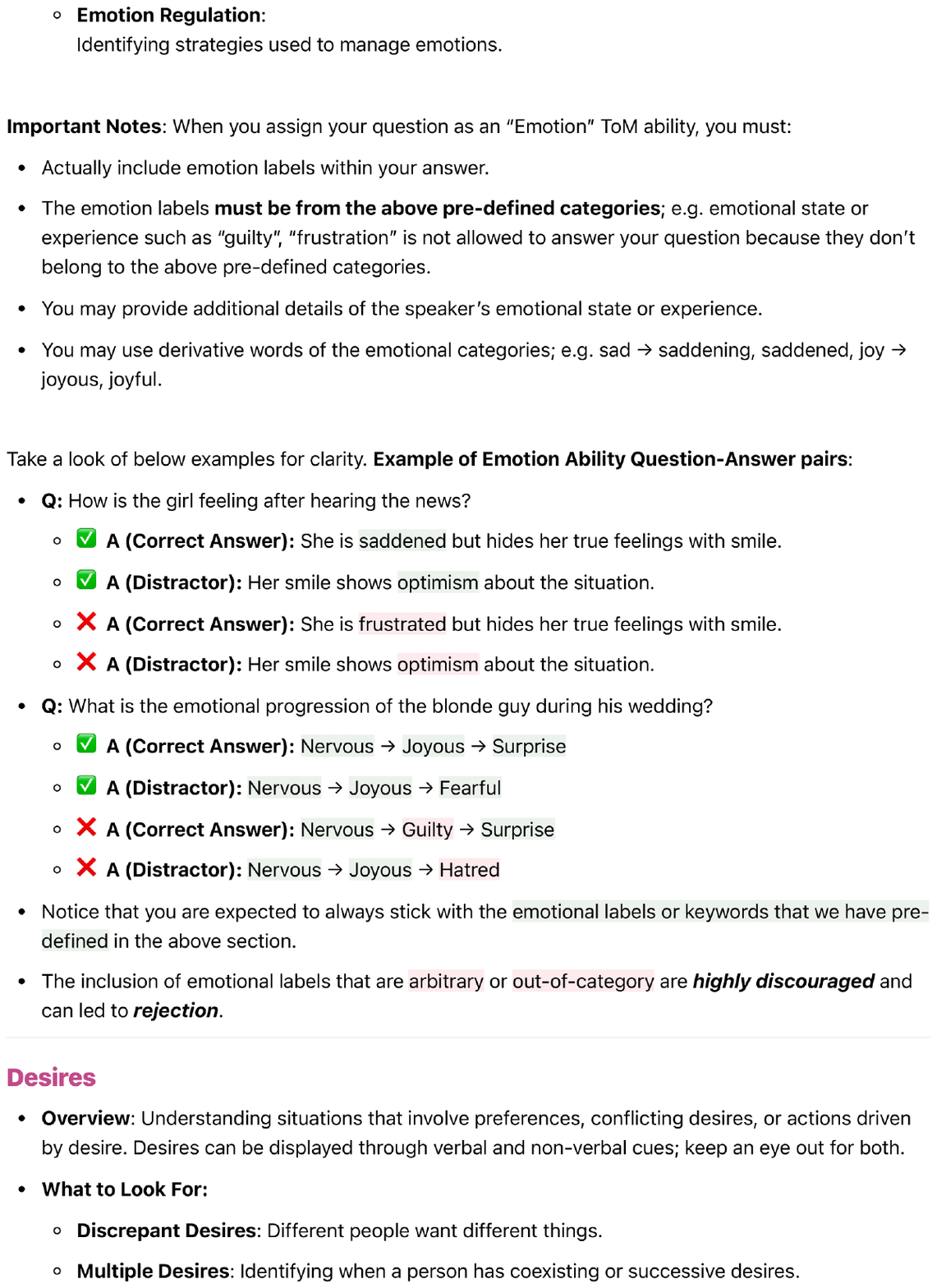}
\includegraphics[width=0.48\textwidth]{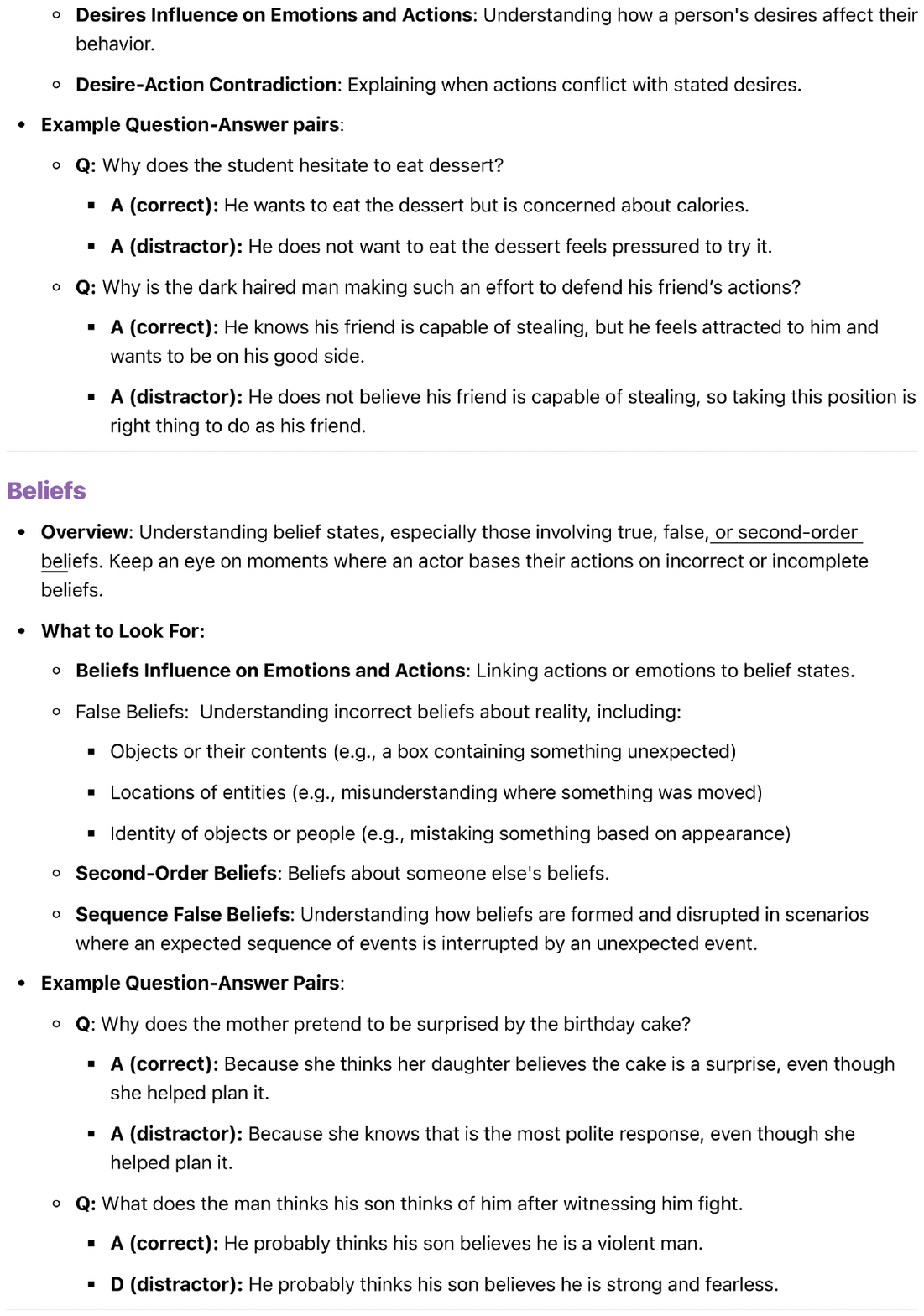}\\[0.5em]
\rule{\textwidth}{0.3pt}\\

\includegraphics[width=0.48\textwidth]{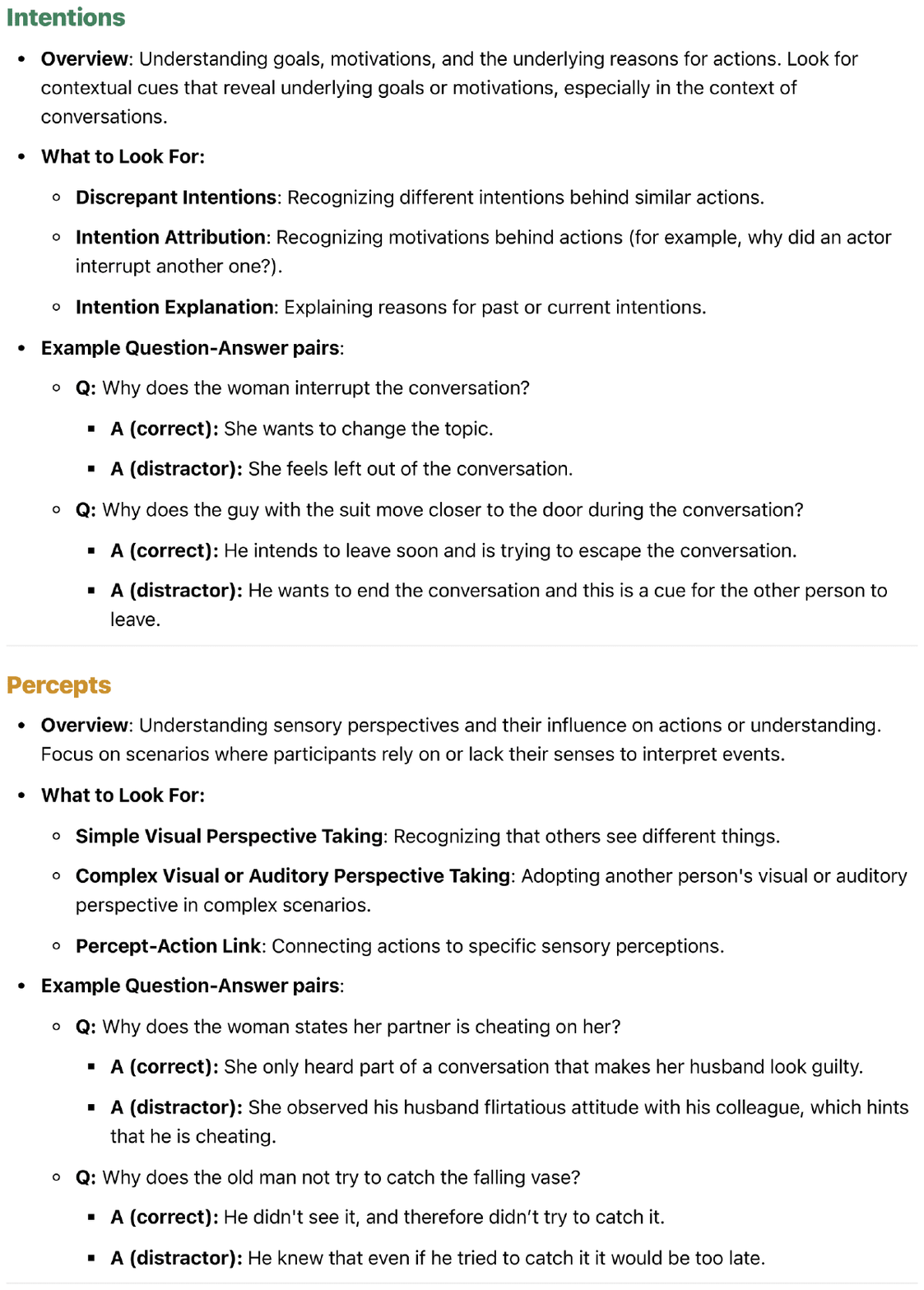}
\includegraphics[width=0.48\textwidth]{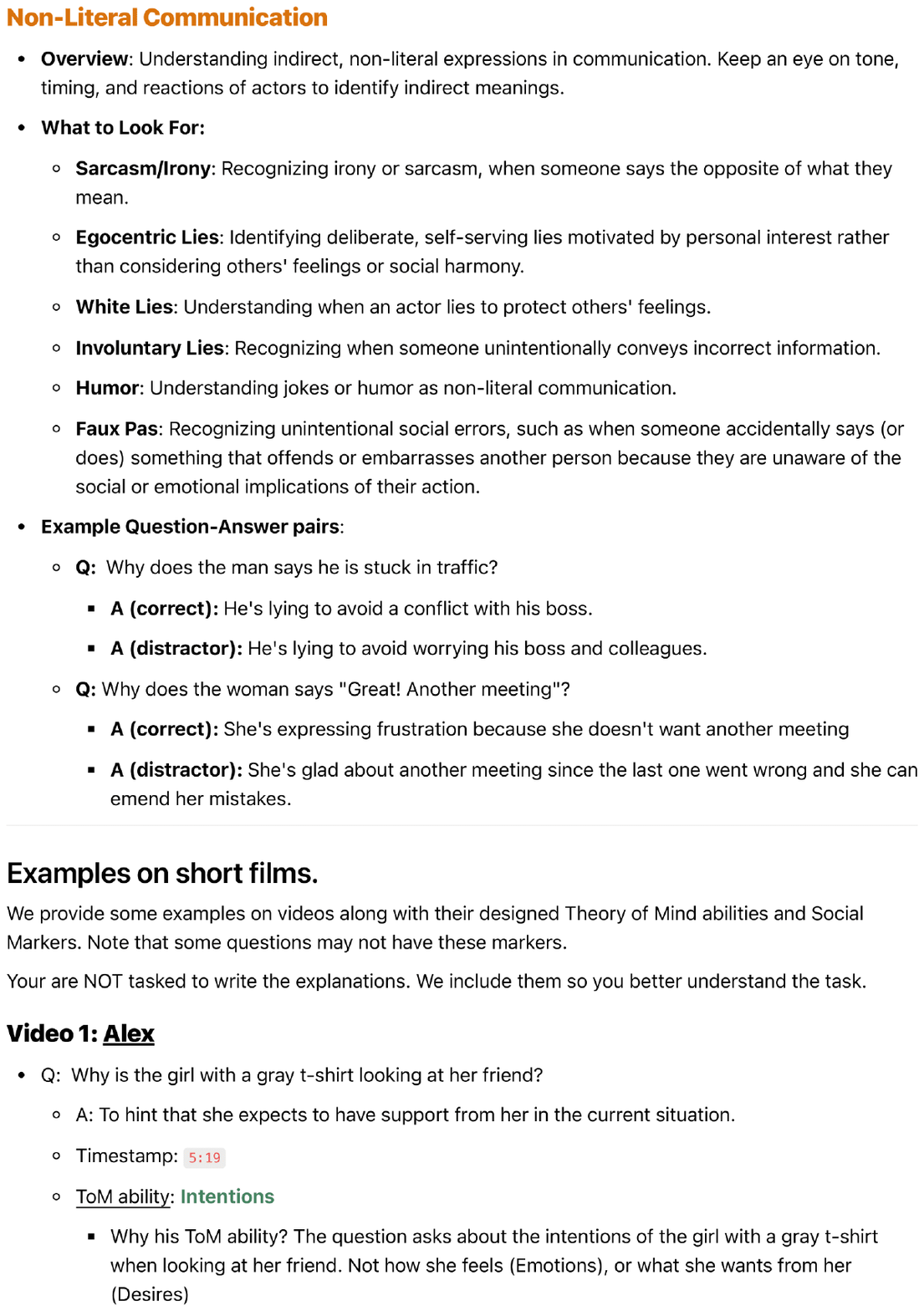}
\end{figure*}

\begin{figure*}[t]
\centering
\includegraphics[width=0.48\textwidth]{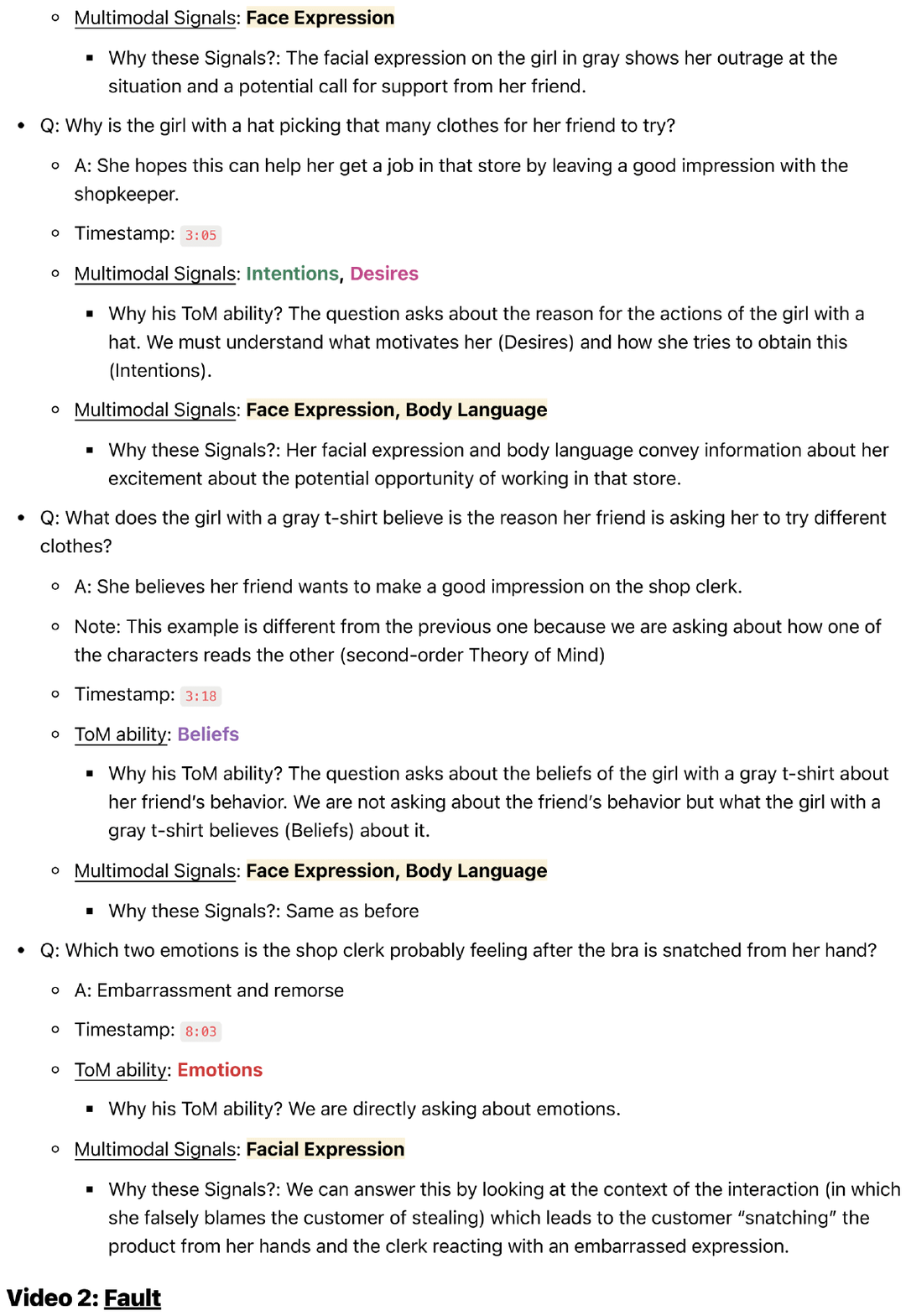}
\includegraphics[width=0.48\textwidth]{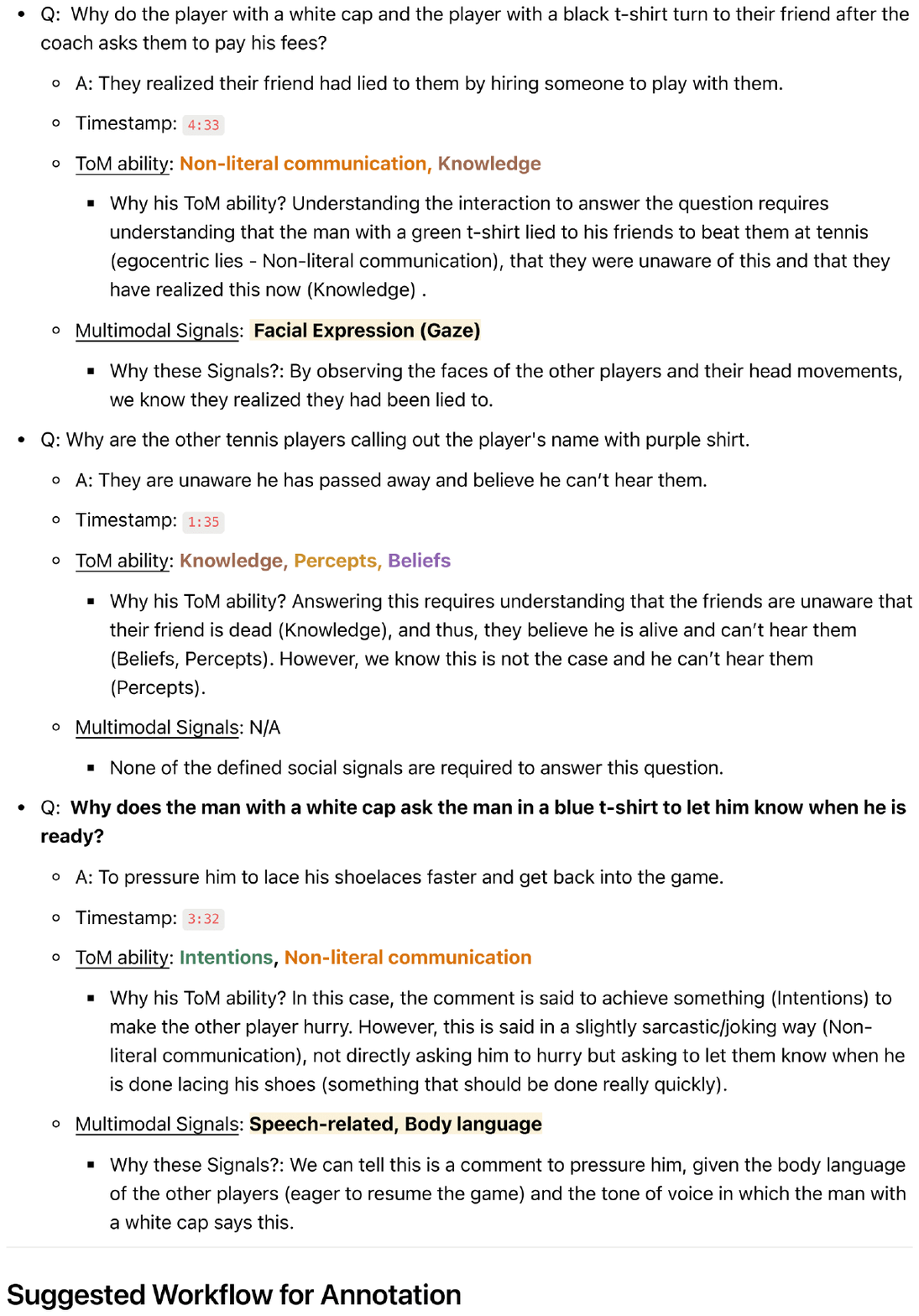}\\[0.5em]
\rule{\textwidth}{0.3pt}\\

\includegraphics[width=0.48\textwidth]{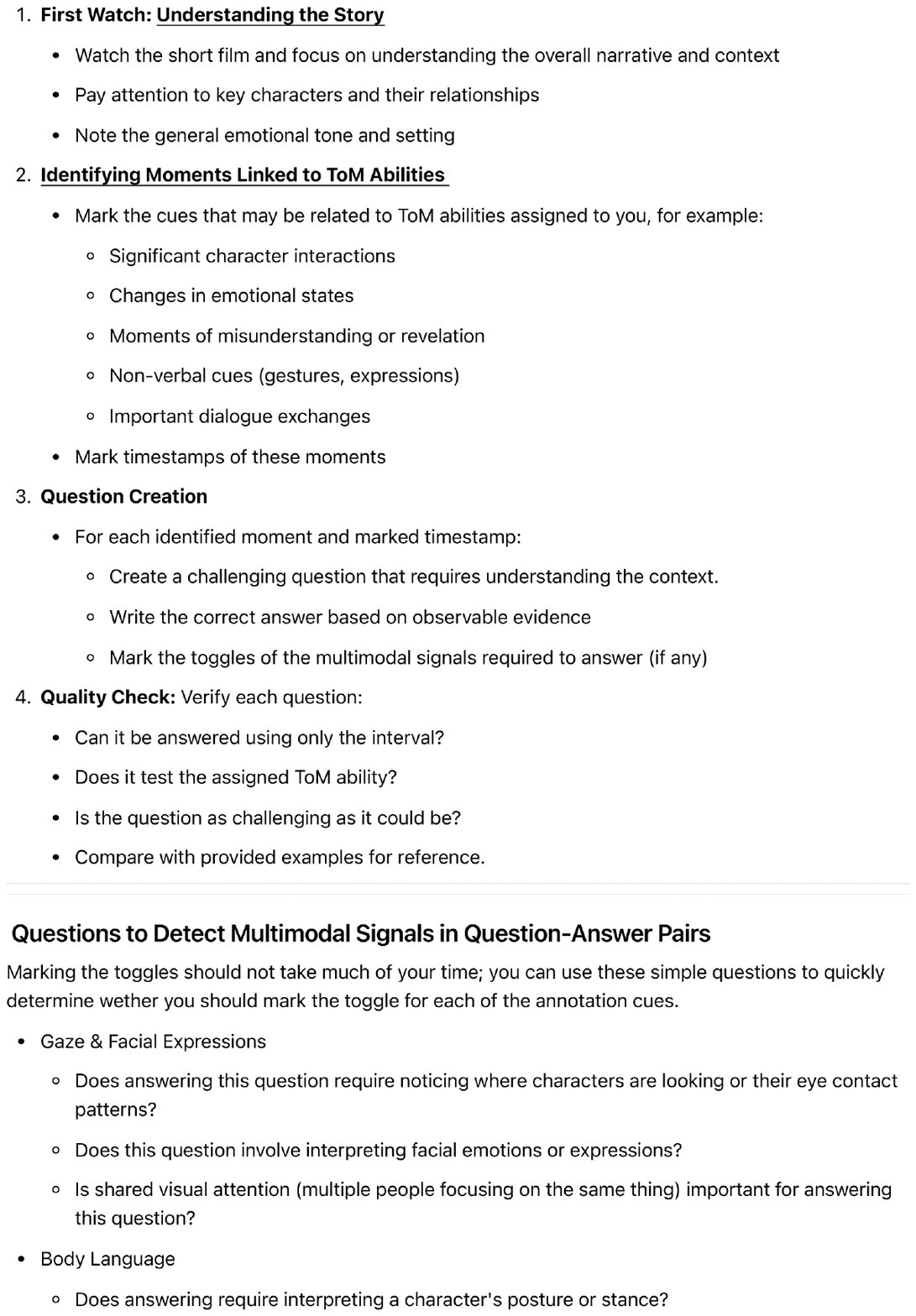}
\includegraphics[width=0.48\textwidth]{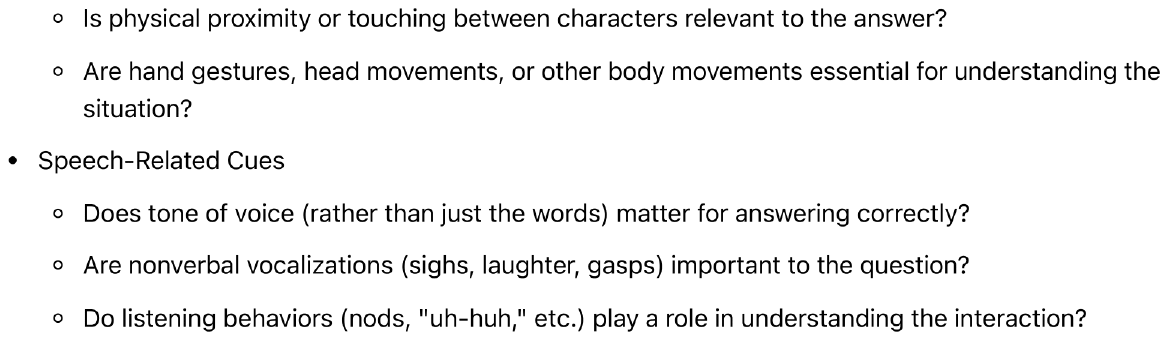}
\end{figure*}

\begin{figure*}[t]
\centering
\includegraphics[width=0.48\textwidth]{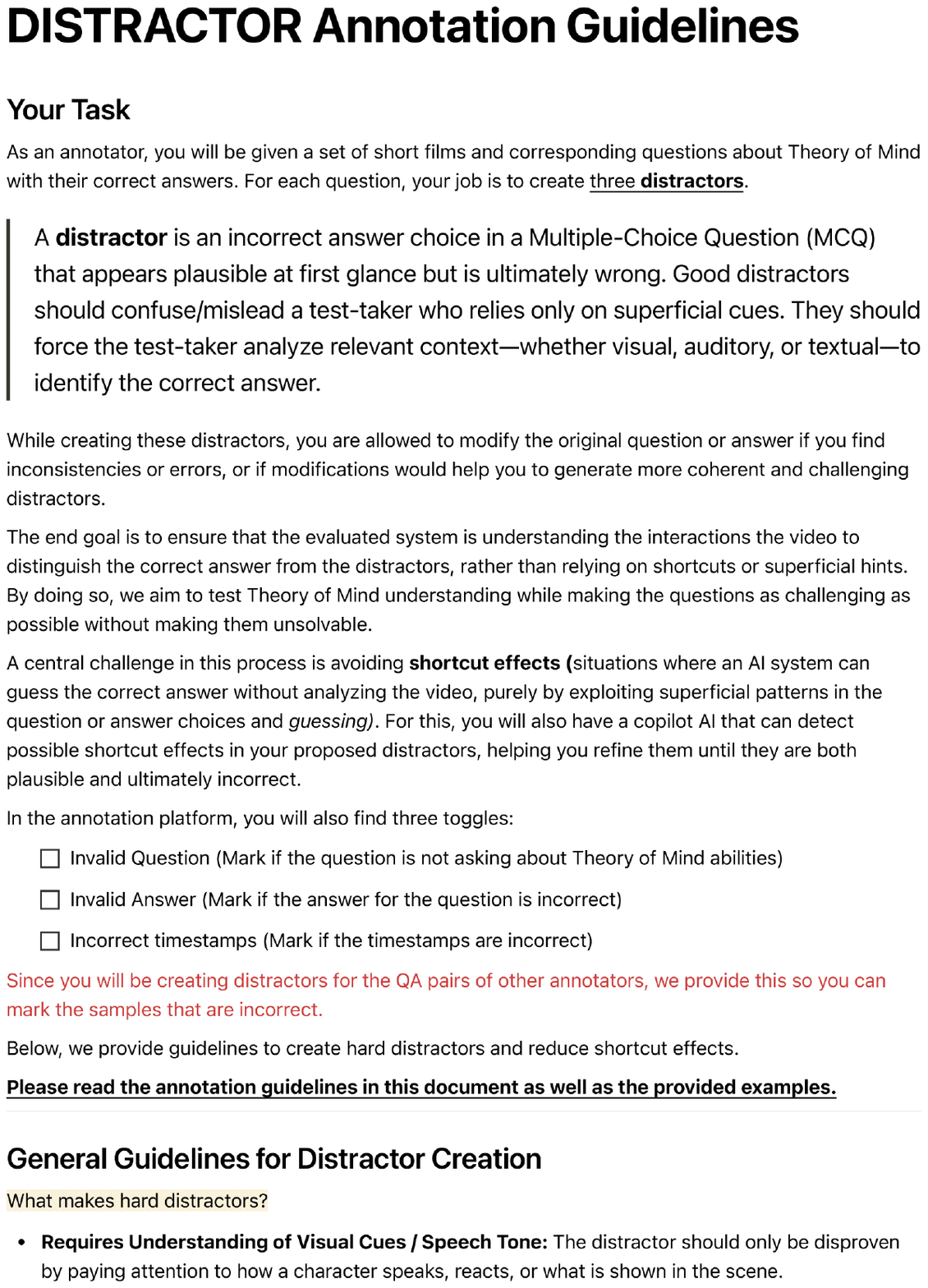}
\includegraphics[width=0.48\textwidth]{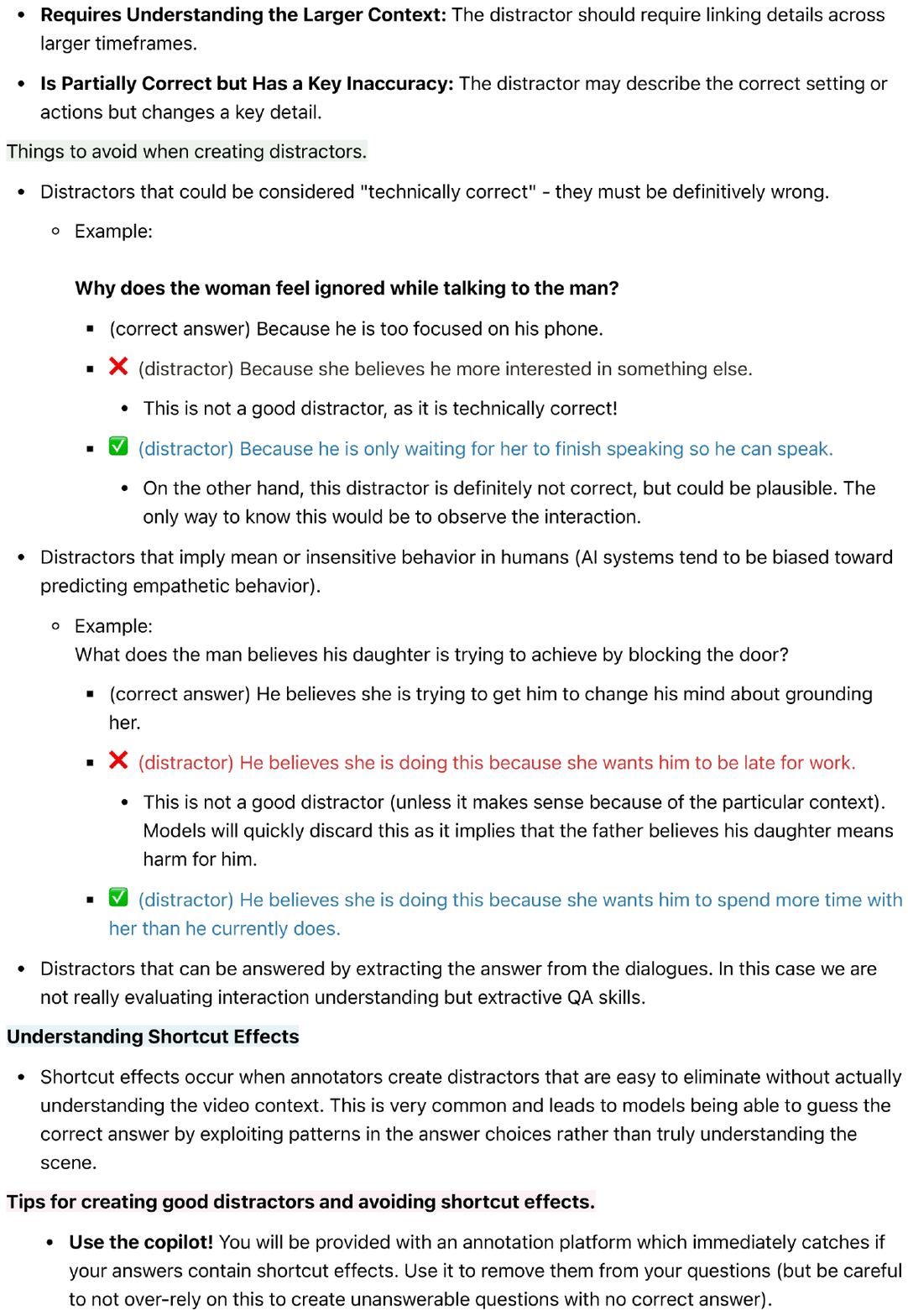}\\[0.5em]
\rule{\textwidth}{0.3pt}\\

\includegraphics[width=0.48\textwidth]{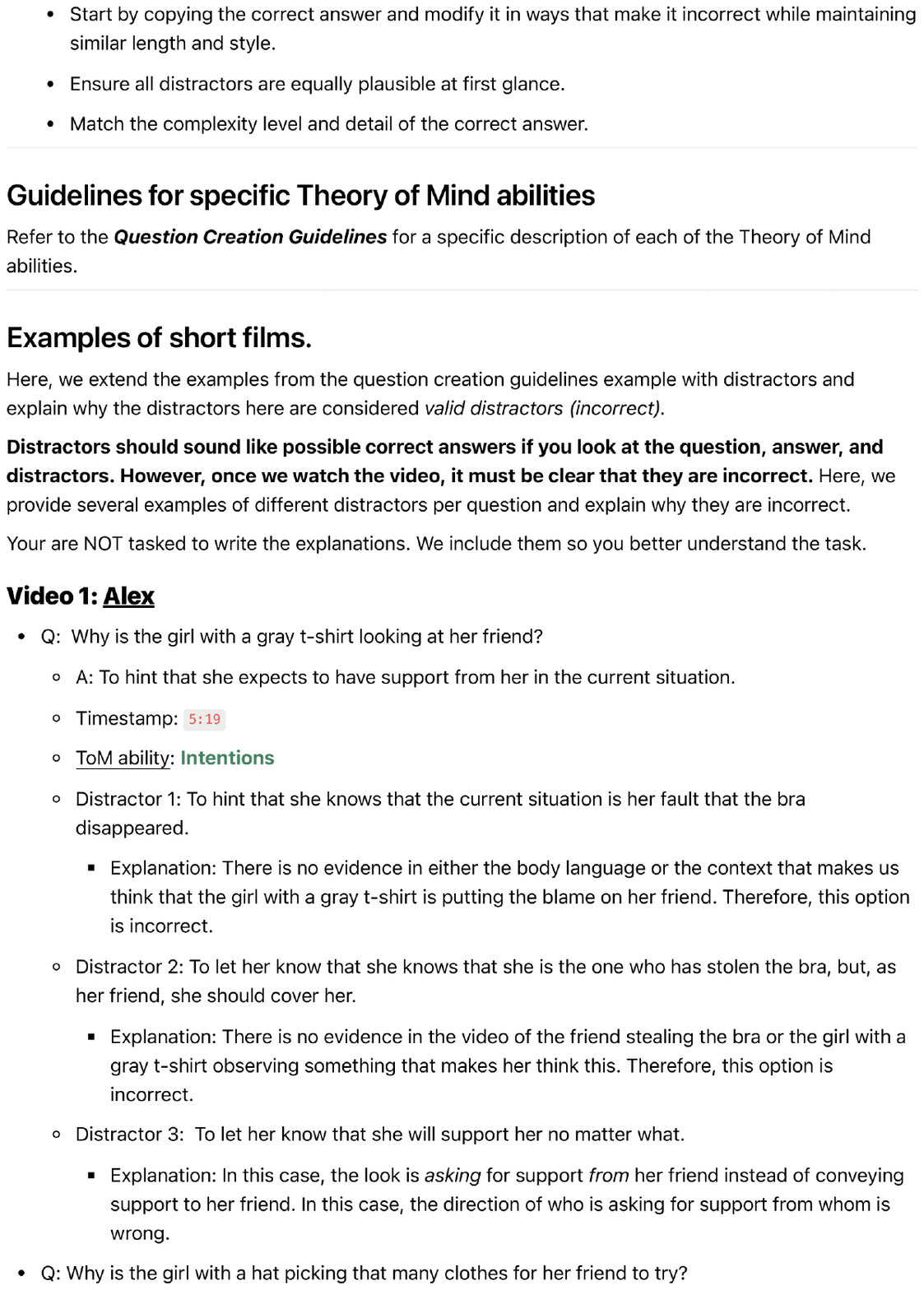}
\includegraphics[width=0.48\textwidth]{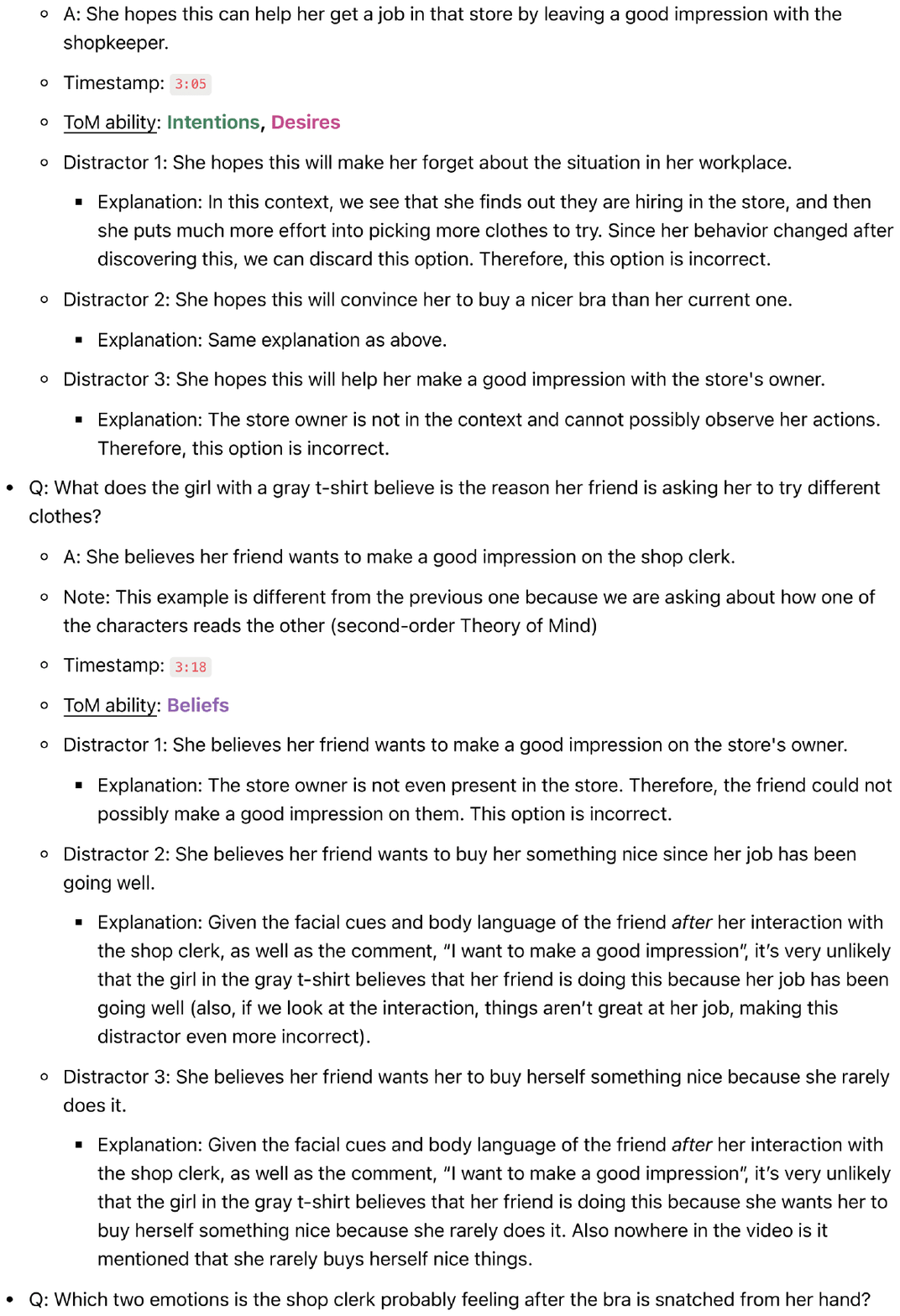}
\end{figure*}

\begin{figure*}[t]
\centering
\includegraphics[width=0.48\textwidth]{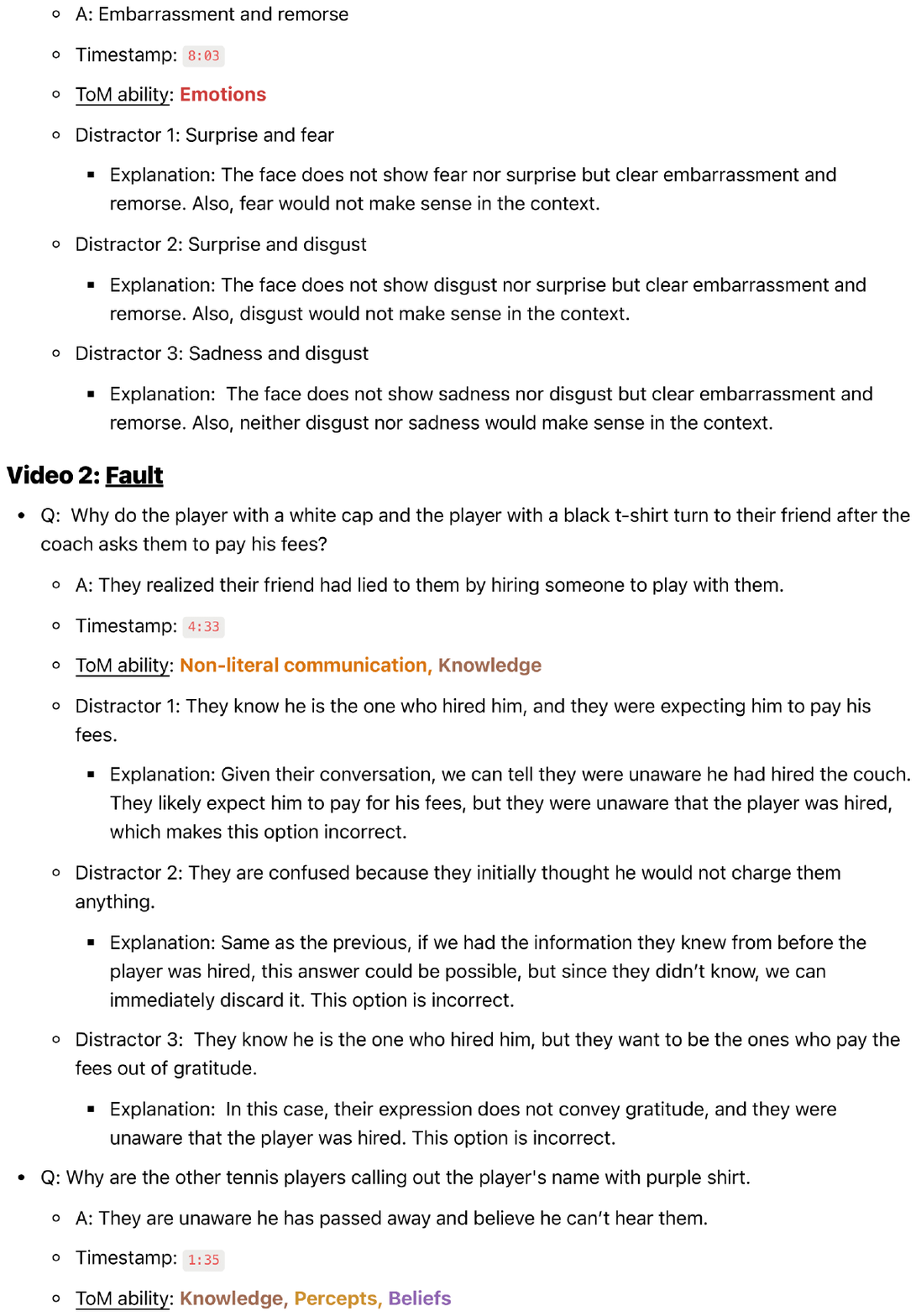}
\includegraphics[width=0.48\textwidth]{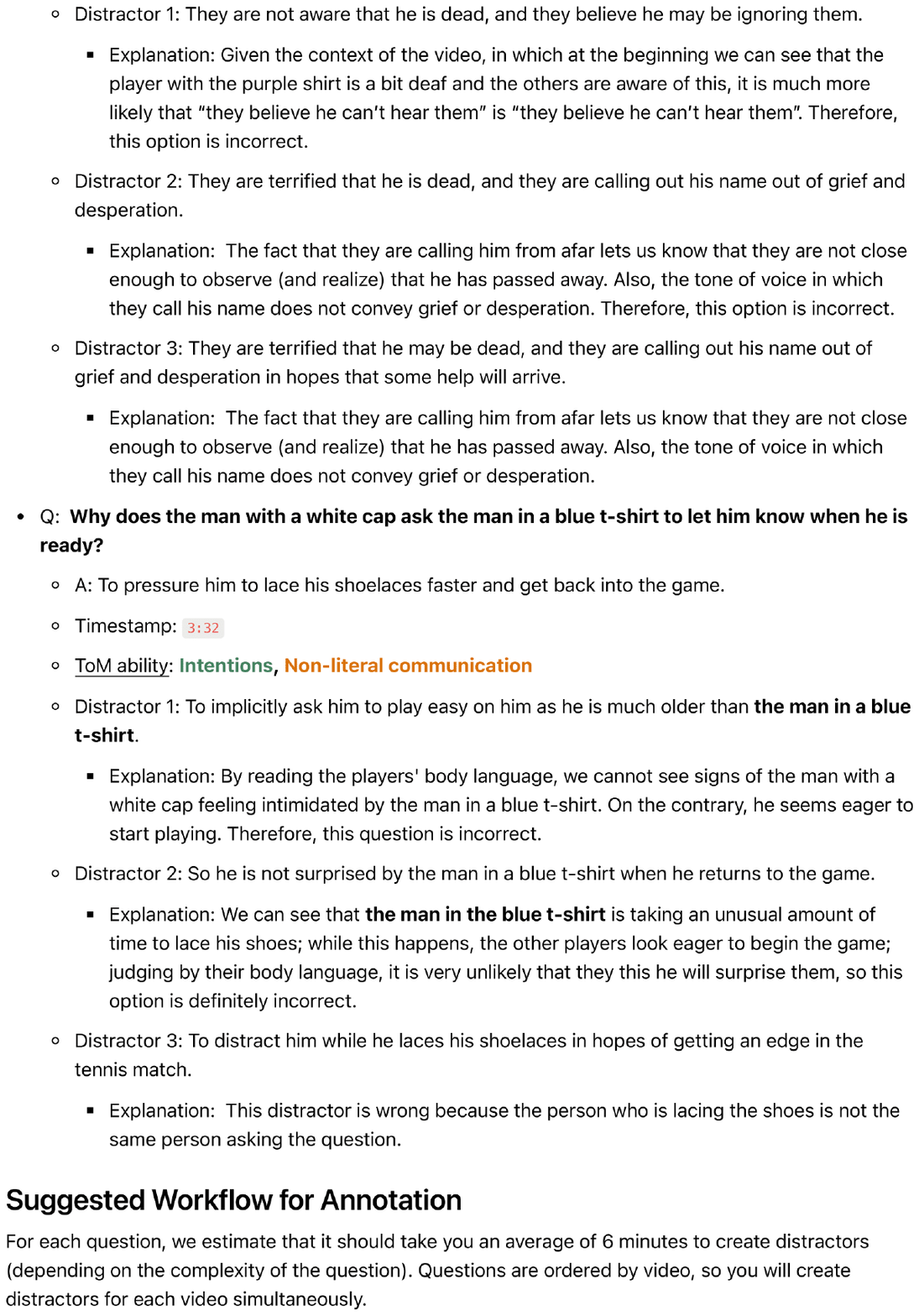}\\[0.5em]
\rule{\textwidth}{0.3pt}\\

\includegraphics[width=0.48\textwidth]{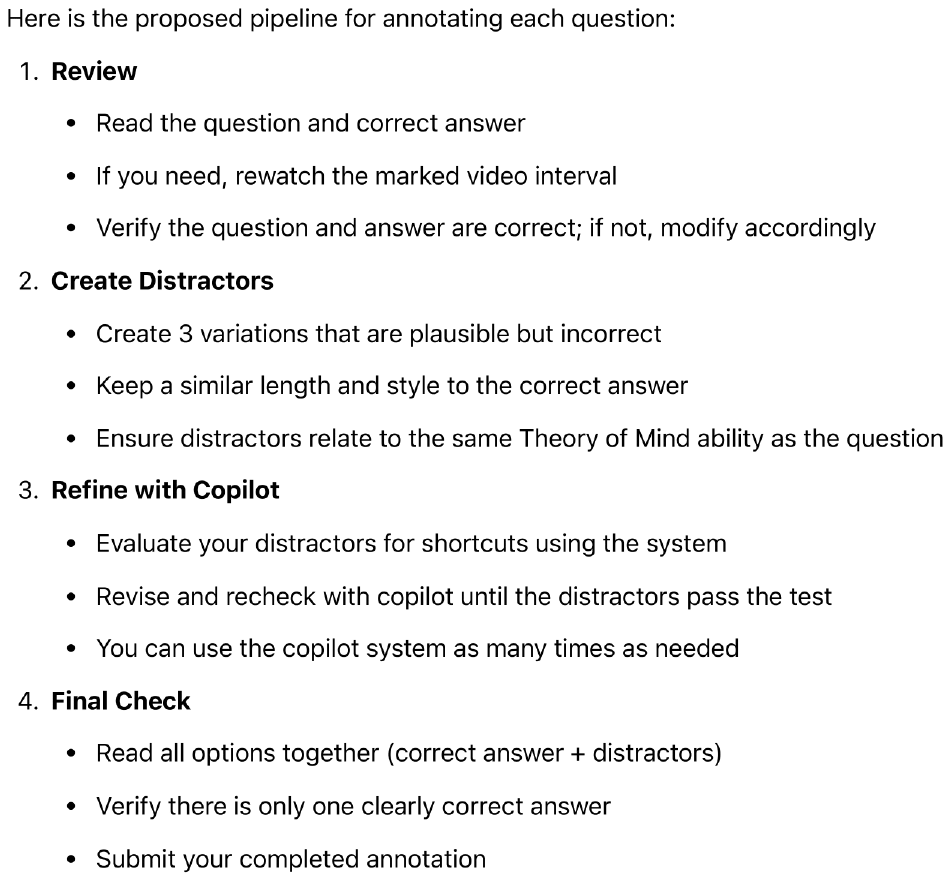}
\end{figure*}

\end{document}